\documentclass[10pt,twocolumn,letterpaper]{article}

\usepackage{cvpr}              

\usepackage{graphicx}
\usepackage{amsmath}
\usepackage{amssymb}
\usepackage{booktabs}
\usepackage{color,colortbl}
\definecolor{Gray}{gray}{0.9}
\definecolor{Gray2}{gray}{0.95}

\usepackage[accsupp]{axessibility}
\usepackage{booktabs}
\usepackage{amsmath}
\usepackage{amssymb}
\usepackage{makecell}
\usepackage{bbm}
\usepackage{xcolor,soul}
\usepackage[accsupp]{axessibility}

\usepackage[pagebackref,breaklinks,colorlinks]{hyperref}

\usepackage[capitalize]{cleveref}
\crefname{section}{Sec.}{Secs.}
\Crefname{section}{Section}{Sections}
\Crefname{table}{Table}{Tables}
\crefname{table}{Tab.}{Tabs.}

\def\cvprPaperID{8224} 
\def\confName{CVPR}
\def\confYear{2022}

\begin{document}

\title{VisualGPT: Data-efficient Adaptation of Pretrained Language Models for Image Captioning}

\author{Jun Chen\textsuperscript{\rm 1} , Han Guo\textsuperscript{\rm 2}, Kai Yi \textsuperscript{\rm 1}, Boyang Li\textsuperscript{\rm 3}, Mohamed Elhoseiny\textsuperscript{\rm 1}\\
\textsuperscript{\rm 1}{ King Abdullah University of Science and Technology (KAUST)}, \\ 
\textsuperscript{\rm 2}{Carnegie Mellon University}, 
\textsuperscript{\rm 3}{ Nanyang Technological University}\\
  \texttt{\{jun.chen,kai.yi,mohamed.elhoseiny\}@kaust.edu.sa} \\ 
  \texttt{hanguo@cs.cmu.edu}, \texttt{boyang.li@ntu.edu.sg} \\
}


\maketitle

\begin{abstract}
The limited availability of annotated data often hinders real-world applications of machine learning. To efficiently learn from small quantities of multimodal data, we leverage the linguistic knowledge from a large pre-trained language model (PLM) and quickly adapt it to new domains of image captioning. To effectively utilize a pretrained model, it is critical to balance the visual input and prior linguistic knowledge from pretraining. We propose VisualGPT, which employs a novel self-resurrecting encoder-decoder attention mechanism to quickly adapt the PLM with a small amount of in-domain image-text data. The proposed self-resurrecting activation unit produces sparse activations that prevent accidental overwriting of linguistic knowledge. When trained on 0.1\%, 0.5\% and 1\% of the respective training sets, VisualGPT surpasses the best baseline by up to 10.0\% CIDEr on MS COCO \cite{lin2014microsoft} and 17.9\% CIDEr on Conceptual Captions \cite{sharma2018conceptual}. Furthermore, VisualGPT achieves the state-of-the-art result on IU X-ray \cite{iuxrayDataset}, a medical report generation dataset. Our code is available at \url{ https://github.com/Vision-CAIR/VisualGPT}.
\end{abstract}


\section{Introduction}


 Recent performance gains in image captioning \cite{kulkarni2013babytalk,xu2015show,karpathy2015deep,cornia2020meshed,huang2019attention} are achieved on top of large-scale data corpora such as MS COCO \cite{lin2014microsoft} or Conceptual Captions~\cite{sharma2018conceptual}, each containing hundreds of thousands of captions. Manual annotation of captions requires considerable time and effort. On the other hand, semi-automatic collection of image-caption pairs from the Internet, as used by Conceptual Captions \cite{sharma2018conceptual}, may generate incorrect or undesirable training data even after multiple rounds of cleaning. Data for specialized domains like medical report generation \cite{YuanLi2018:MedicalReport,iuxrayDataset} and low-resource language captioning \cite{Wu:low-resource-2019,arabic} cannot be easily scaled. Improving the data efficiency of image captioning networks would enable quick data curation, description of rare objects, and applications in specialized domains. 





\begin{figure}
\centering
 \includegraphics[width=0.8\linewidth]
                  {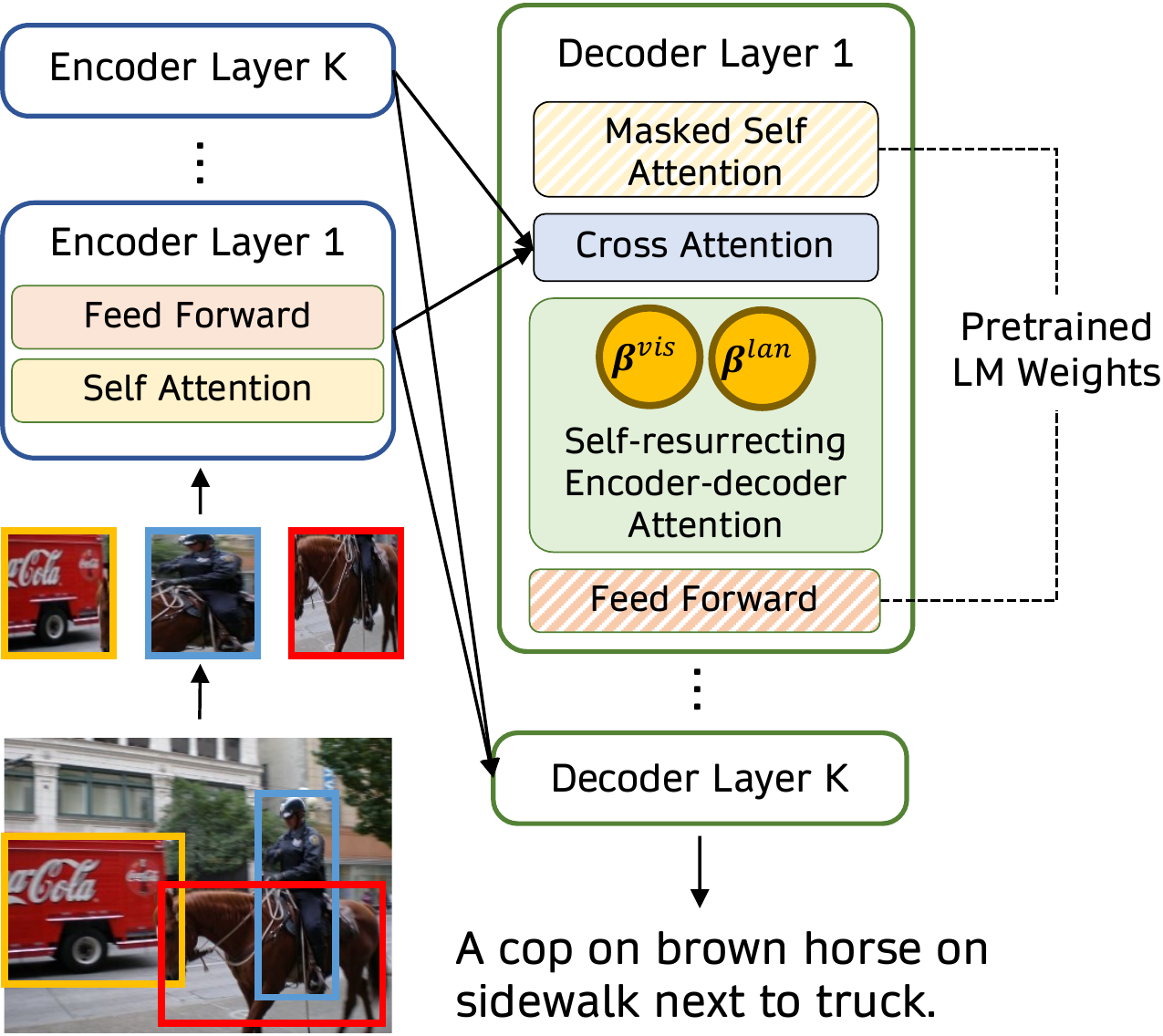}
\caption{Our VisualGPT model transfers the knowledge from a pre-trained language model to the caption decoder. A self-resurrecting encoder-decoder attention is designed to connect the multi-level visual features and caption decoder.}
\label{fig:teaser}
\end{figure}



%

In this paper, we investigate the data efficiency problem for image captioning. This problem is distinct from the novel object captioning problem \cite{hendricks2016deep,agrawal2019nocaps}, which relies on abundant in-domain data but zero out-of-domain data. Instead, we aim to improve the performance of image captioning systems trained on a small subset of \emph{ in-domain} data.

We propose to improve data efficiency by leveraging pre-trained language models (PLMs) \cite{dong2019unified,liu2019roberta,lewis-etal-2020-bart,raffel2020exploring}, such as BERT \cite{devlin2018bert}, XLNet \cite{yang2019xlnet}, and GPT \cite{gpt1,radford2019language,brown2020language}. Via self-supervised learning, these models acquire rich linguistic and semantic knowledge, which has been shown to inform downstream tasks in NLP \cite{hellogpt,golovanov2019large}. However, the adaptation of PLMs pretrained on unimodal textual data for multimodal tasks remain under-investigated. 

\begin{figure}
\centering
 \includegraphics[width=0.75\linewidth]
                  {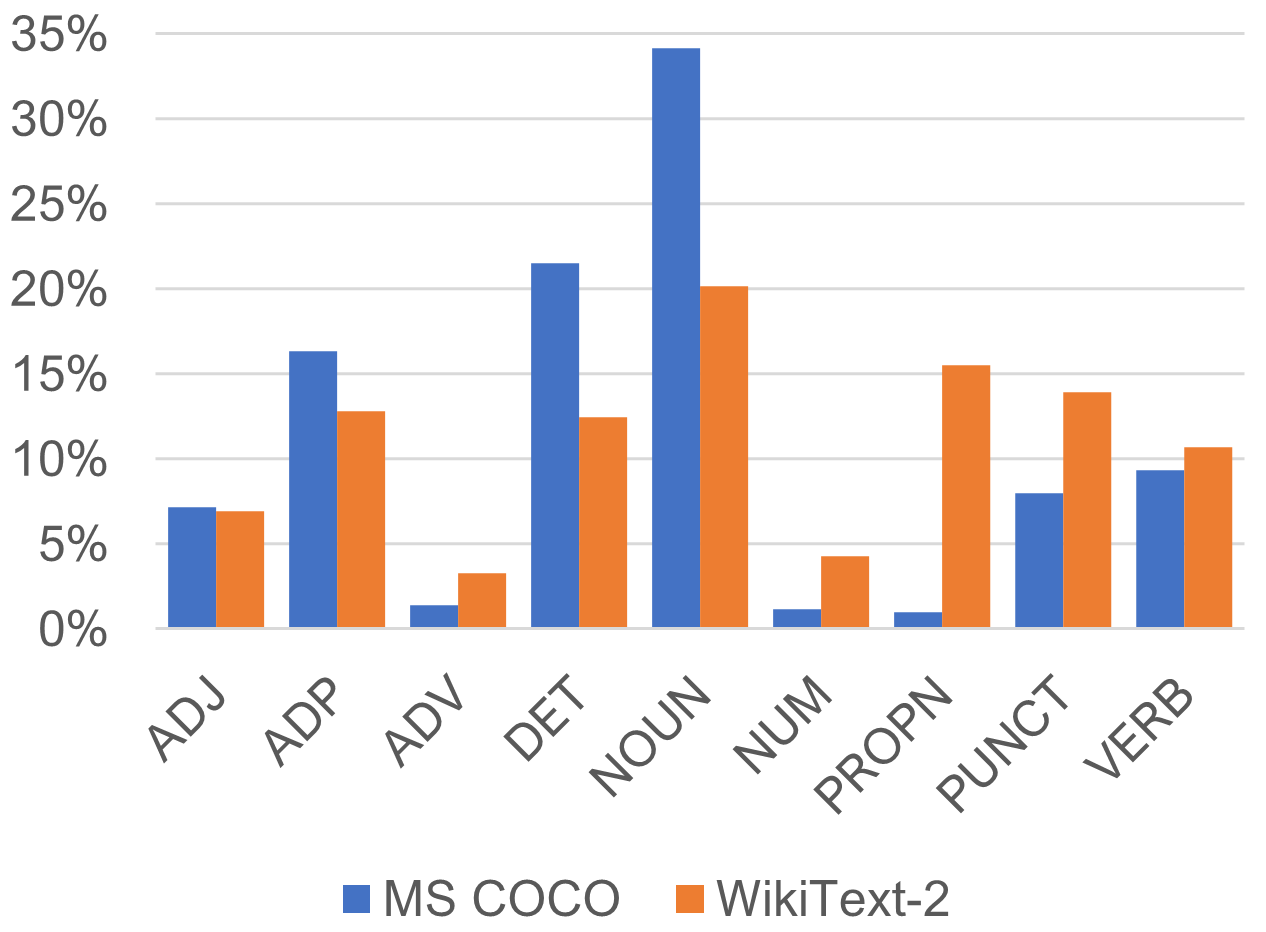}
\caption{Comparison of the part-of-speech distributions of the MS COCO and WikiText-2 datasets \cite{Merity2017:wikitext}. We use the spacy parser and show only the most important categories. }
\label{fig:pos-comparison}
\end{figure}

A key challenge in utilizing PLMs is to bridge the domain gap between multi-modal data and the unimodal textual data the PLMs are pre-trained on. In Figure \ref{fig:pos-comparison}, we compare the part-of-speech distributions of MS COCO and WikiText-2 \cite{Merity2017:wikitext}. MS COCO employs $75\%$ more nouns but $14\%$ fewer verbs, which indicates a bias toward descriptions of static objects rather than actions. This suggests that, in order to effectively utilize PLMs in image captioning, we must balance prior linguistic knowledge acquired from pretraining and visual input information.

Figure \ref{fig:teaser} depicts the overall architecture of our proposed model, dubbed as VisualGPT. In the commonly used encoder-decoder architecture for image captioning, we initialize the parameters of the decoder from PLMs such as GPT-2 \cite{radford2019language}, whereas the encoder layers are randomly initialized. In addition, we propose an attention mechanism with self-resurrecting activation units (SRAUs), which balances the input from the visual encoder and the linguistic input from the previous decoder layer. 
The proposed mechanism can produce sparse activations while not being as vulnerable to the zero-gradient problem as regular gates; the self-resurrecting gates can be ``turned on'' again after being zeroed out.

Empirical results demonstrate that, when trained on $0.1\%$, $0.5\%$, and $1\%$ of the MS COCO and Conceptual Captions data, VisualGPT outperforms several strong baseline models. We achieve the state-of-the-art result on IU X-ray \cite{iuxrayDataset}, a medical report generation dataset. With several ablation experiments, we verify the effectiveness of PLMs and the proposed self-resurrecting attention mechanism.

\vspace{0.1in}
\noindent \textbf{Contributions.} We make the following contributions:
\begin{itemize}
\item We explore the data efficiency problem for image captioning by utilizing pretrained language models (PLMs) as the caption decoder. With only a small amount of in-domain training data, the proposed technique quickly adapts PLMs to the cross-modal task of image captioning. To our knowledge, this is the first work that focuses on efficiently adapting large pretrained language models for image captioning.
    
\item We propose a novel encoder-decoder attention with self-resurrecting activation units (SRAUs), which can balance features from the visual and textual modalities. SRAU produces sparse activations that reduce accidental overwriting of pretrained weights. 
\end{itemize}





\section{Related Work}
\label{sec:related-work}

\noindent \textbf{Image Captioning.} Image captioning has been extensively studied in computer vision research. Early methods \cite{farhadi2010every,li2011composing,socher2010connecting,yao2010i2t,kulkarni2013babytalk} focus on filling templates with extracted objects, attributes, and relationships. With the advent of deep learning, researchers proposed end-to-end neural networks that encode an image into vector representations and decode a caption word by word \cite{vinyals2016show,johnson2016densecap}. Many improvements to the encoder \cite{chunseong2017attend,xu2015show,lu2017knowing,yao2018exploring, yang2019auto,yao2019hierarchy,li2019know}, the decoder \cite{YangZhilin:ReviewNetwork2016,wang2017skeleton,wang2018cnncnn}, and the attention mechanism \cite{chen2017sca,lee2018stacked,li2019entangled, huang2019attention,cornia2020meshed} has since been proposed. Encoding the image using object regions has proven beneficial \cite{anderson2018bottom}. Reinforcement learning enables model optimization with non-differentiable evaluation metrics \cite{self_critical,policy_gradient,dai2017towards,shetty2017speaking}. 
\cite{cornia2019show,chen2020say} investigate fine-grained control of caption generation. \cite{dai2017towards,shetty2017speaking} adopt GAN-like architectures that encourage human-like captions. 

A few formulations of the image captioning problem deviate from the traditional supervised learning paradigm. 
Novel object captioning aims to describe objects that do not exist in the training data \cite{hendricks2016deep,venugopalan2017captioning,lu2018neural,li2019pointing,agrawal2019nocaps}.
Feng \textit{et al.} \cite{feng2019unsupervised} propose unsupervised captioning without using paired image-caption supervision. 
Kim \textit{et al} \cite{scarceimagecaption} focus on learning efficiency and improve the data efficiency by learning from auxiliary unpaired image-caption data.




\begin{figure*}
\centering
 \includegraphics[width=1.0\linewidth]
                  {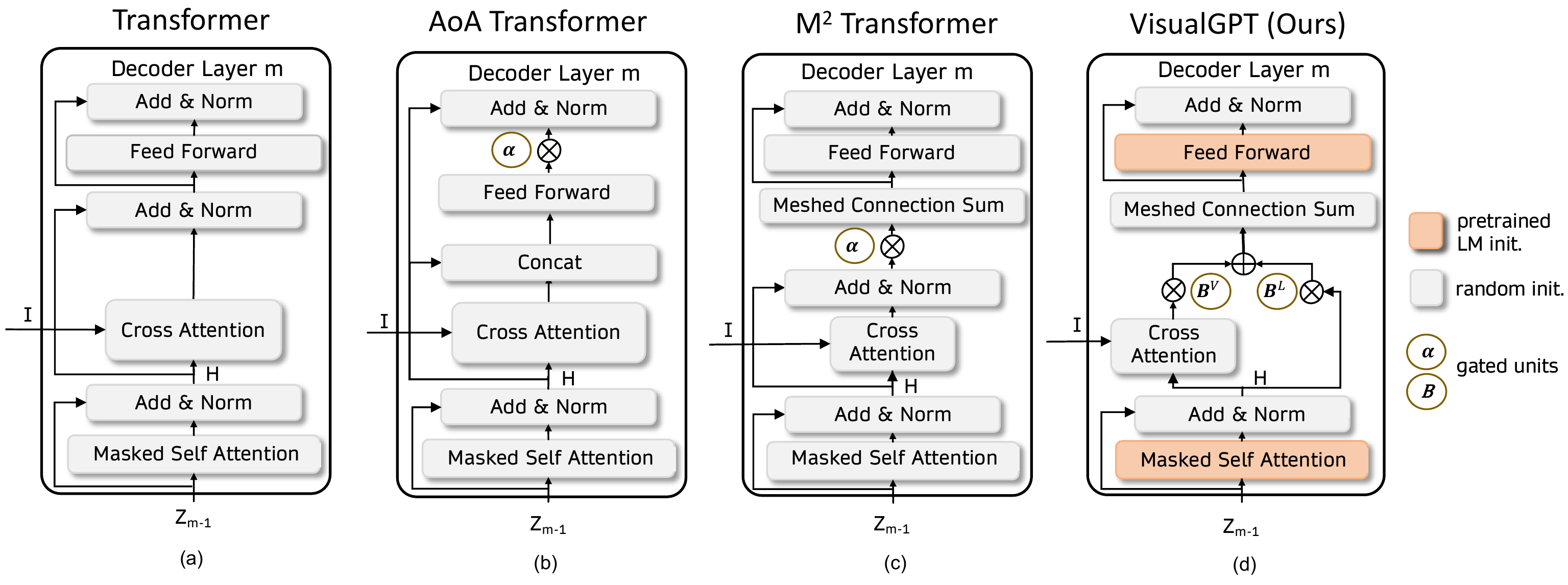}

\caption{Architectures of vanilla Transformer \cite{vaswani2017attention}, Transformer with AoA module \cite{huang2019attention} (AoA Transformer), $\mathcal{M}^{2}$ Transformer \cite{cornia2020meshed}, and VisualGPT. We denote $I$ and $H$ as the visual and language features, respectively. $Z_{m-1}$ is the output from decoder layer $m-1$. Within the circles, $\alpha$, $B^{V}$ and $ B^{L}$ represent different gating units. }
\label{fig:main_model}

\end{figure*}

\vspace{0.1 in}
\noindent \textbf{Self-supervised NLP Models.} Self-supervised training of large neural networks on textual data proves to be an important technique in the creation of high-performance NLP models. Several self-supervision signals have been proposed, such as autoregressive language modeling \cite{bengio2003neural,mikolov2011extensions}, which includes the GPT series of models \cite{radford2019language,brown2020language,gpt1}, and masked language modeling, which includes ELMo \cite{elmo} and BERT-related methods \cite{devlin2018bert, albert,roberta}. 



In this paper, we propose a quick adaptation technique for network weights obtained using the language modeling (LM) objective. However, the proposed technique can easily be applied to other models, as the masked language modeling objective can be converted to the LM objective by masking only the last word in the textual sequence. 
Unlike neural networks pretrained on multimodal data (e.g., \cite{qi2020imagebert,tan2019lxmert,su2019vl,vilbert,oscar,vinvl,zhuge2021kaleido}), our method only requires a small amount of multimodal training data and focuses on adapting linguistic knowledge learned from the textual modality. 


\section{Preliminaries: Transformer for Captioning}
The Transformer \cite{vaswani2017attention} has become one of the standard models for image captioning. At its core lies the multi-head dot-product attention mechanism.  Taking three input matrices, query $Q$, key $K$, and value $V$, the attention function can be written as
\begin{equation} \text{Attn}(Q,K,V) =\text{softmax}\left( \frac{(W^q Q) (W^k K)^{\top}}{\sqrt{D}}\right) W^v V,
\label{eq:fatt}  
\end{equation}
where $W^q, W^k$, and $W^v$ are trainable parameters and D is a scaling factor. Intuitively, the attention operation can be seen as encoding $W^qQ$ as convex combination of the row vectors of $W^vV$. 
The multi-head attention repeats the process with multiple sets of $W^q, W^k$, and $W^v$; the results are concatenated and linearly projected back to the same dimensionality. 

In visual captioning tasks, we apply a visual encoder whose output is $I \in \mathbb{R}^{O\times S}$. $O$ is the length of the input sequence, which in this work is a sequence of objects in the image. $S$ is the hidden dimension size. The decoder network outputs words in the caption sequentially. 

When decoding word $t+1$, the encoder-decoder attention takes as input the visual encoding $I$ and the current state of the decoder $H \in \mathbb{R}^{t \times S}$. We apply the attention operation with $H$ as the query and $I$ as both the key and the value. The encoder-decoder attention is then
\begin{equation} \text{EncDecAttn}(H, I) = \text{Attn}(H, I, I).
\end{equation}
After that, we apply the AddNorm operator, which contains a residual connection and layer normalization \cite{ba2016layer} and can be written as
$\text{LayerNorm}(\text{EncDecAttn}(H, I) + H)$.

Researchers have proposed other variants of the encoder-decoder attention. In Figure \ref{fig:main_model}, we contrast these decoder architectures with the proposed VisualGPT model. The Attention-on-Attention (AoA) module  \cite{huang2019attention} provides an alternative method for combining the visual encoding $I$ and the linguistic information $H$ from the decoder. 
For another method for combining visual and linguistic information, $\mathcal{M}^{2}$ Transformer \cite{cornia2020meshed} connects all decoder layers to all encoder layers. In Figure \ref{fig:main_model}, it is represented by the box labeled as \emph{Meshed Connection Sum}.

\section{VisualGPT}
Pretrained language models (PLMs) such as GPT-2 \cite{radford2019language} are trained on data from a single modality. We use a PLM as the caption decoder and feed visual information to the PLM via the encoder-decoder attention, which plays a crucial role in quickly adapting the PLMs.


With the design of the encoder-decoder attention, we aim to carefully balance visual information from the encoder and linguistic knowledge stored in the PLM.
During the generation of visual words, such as ``person'', ``truck'', or ``dog'', the model should attend to visual information. In contrast, the generation of determiners or connectives requires only linguistic knowledge. Ideally, we would like to exploit the massive amount of linguistic knowledge stored in the PLM weights (e.g., ~\cite{liu2019linguistic}), while referring to the visual input only when required. To achieve this goal, we introduce a pair of specialized gating units.



\subsection{Self-Resurrecting Activation Unit}
The encoder-decoder attention EncDecAttn$(H, I)$ may be seen as encoding the linguistic information $H$ with visual information $I$. In VisualGPT, we control the balance between these two modalities using two complementary gates $B^{\text{vis}}$ and $B^{\text{lan}}$. The output of this module is
\begin{equation}
B^{\text{vis}} \otimes \text{EncDecAttn}(H, I) + B^{\text{lan}} \otimes H,
\end{equation}
where $\otimes$ denotes element-wise multiplication. Letting $B^{\text{vis}}[i, j]$ and $B^{\text{lan}}[i, j]]$ denote the elements in the matrices, they are computed in pairs as
\begin{equation} 
\label{eq:srau-2}
\begin{split}
B^{\text{vis}}[i, j] & = \sigma(H[i,j]) \mathbbm{1} (\sigma(H[i,j])> \tau) , \\  
B^{\text{lan}}[i, j] & = (1- \sigma(H[i,j]) \mathbbm{1}(1- \sigma(H[i,j])> \tau),
\end{split}
\end{equation}
where $\tau$ is a predefined threshold hyperparameter and $\mathbbm{1}(\cdot)$ is the indicator function, which returns 1 if the inner statement is true and 0 otherwise. 

\begin{figure}[t!]
\centering
\scalebox{1.0}{
 \includegraphics[width=0.5\linewidth]
 {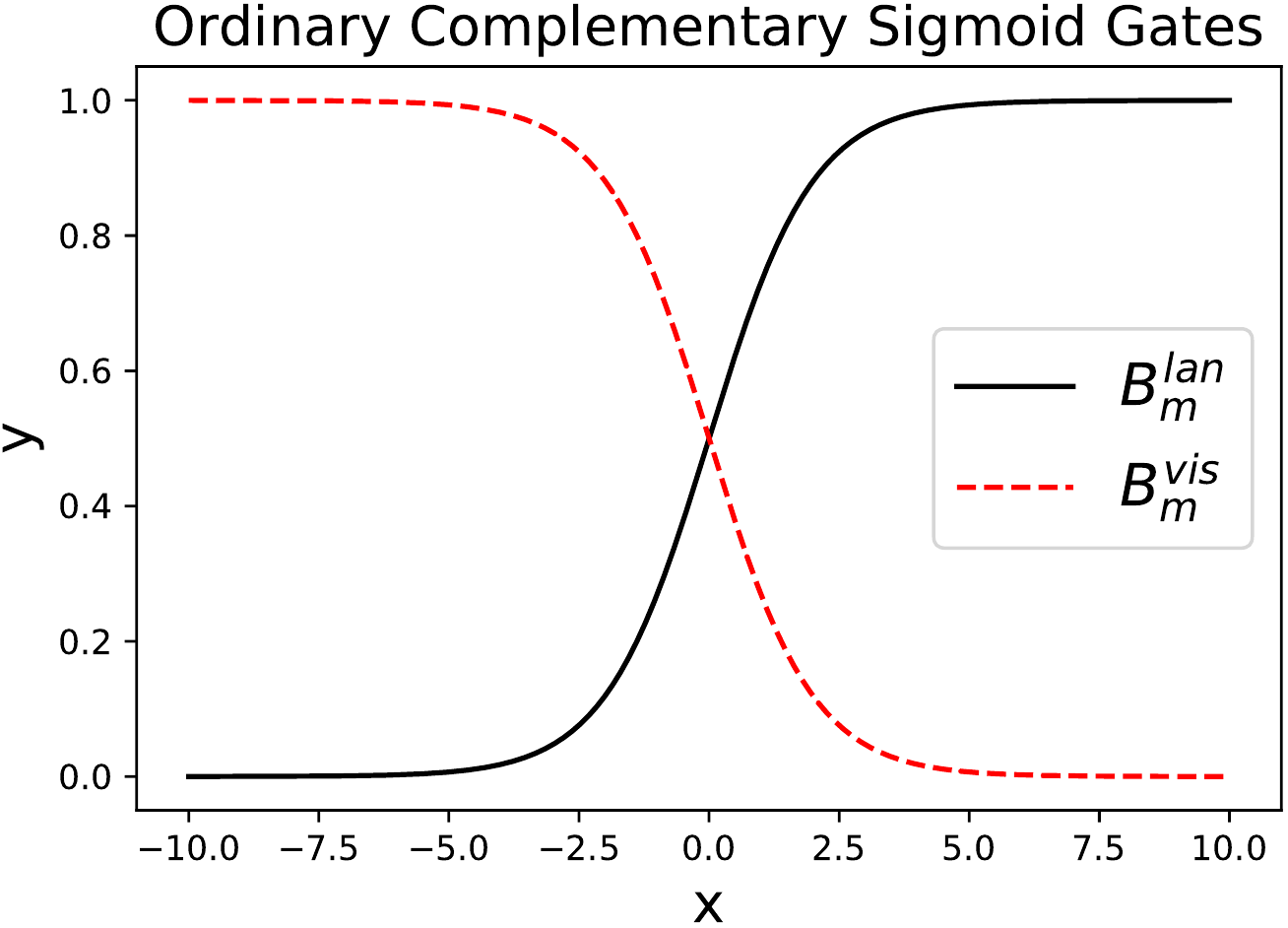}
 \includegraphics[width=0.5\linewidth]
         {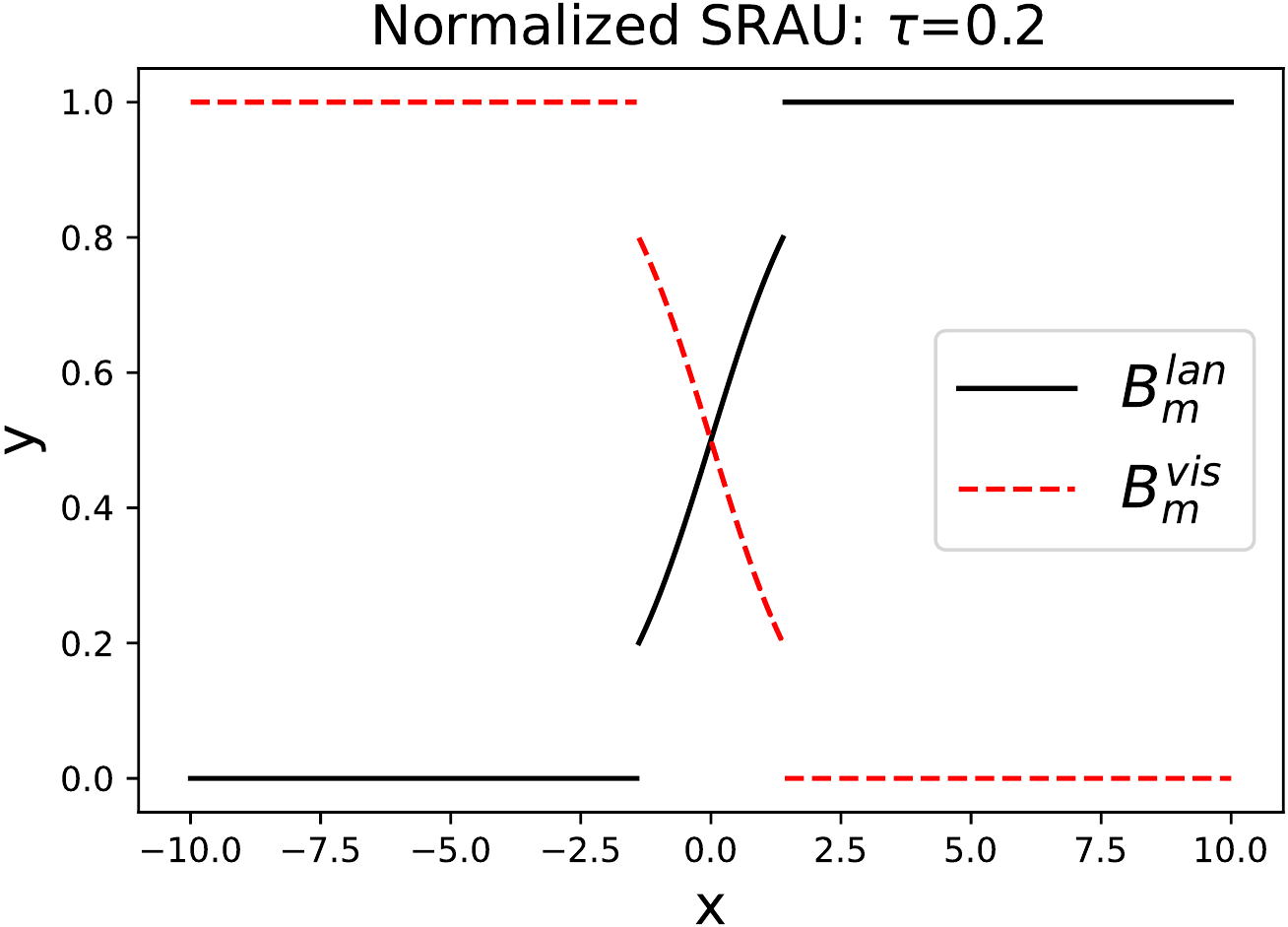}
         }
 \includegraphics[width=0.5\linewidth]
                  {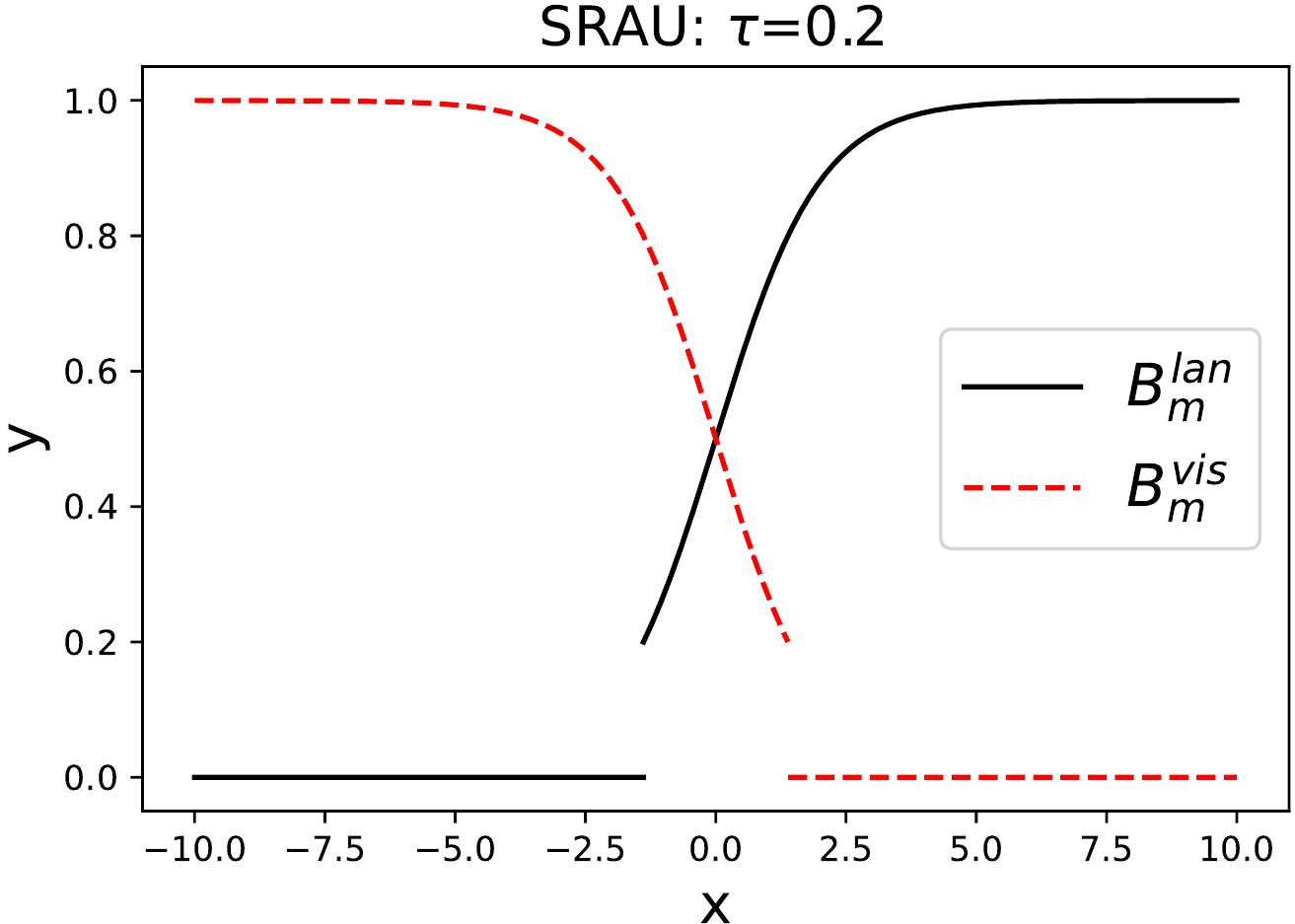}
\caption{Top Left: Ordinary complementary sigmoid gates. Top Right: Normalized SRAU $\tau$=0.2.  Bottom:  SRAU $\tau$=0.2. The x-axis indicates the function inputs and the y-axis indicates outputs.}
\vspace{-0.15in}
\label{fig:SRAU}
\end{figure}

An alternative to SRAU is ordinary complementary gates (OCG), computed as $\sigma(H[i,j])$ and $1-\sigma(H[i,j])$ (see Figure \ref{fig:SRAU}, top left). OCG can output values that are very close to zero. In contrast, with the indicator functions SRAU directly sets values less than the threshold $\tau$ to zero, thereby introducing sparsity.  When $\tau$ is set to 0, SRAU becomes OCG. 
%
%
As the gradient cannot backpropagate through zero gates, SRAU prevents optimization from disrupting pretrained weights that capture linguistic knowledge. This property is crucial in effective utilizing of pretrained models. In contrast, when the OCG gates output near-zero values, some small but non-zero gradients may still overwrite existing linguistic knowledge.

Another advantage of SRAU is its ability to escape from zero outputs. It is possible for one gate to output zero and have zero gradient while the gradient for the other gate remains usable (e.g., when $x$ in Fig 4 is close to $1.3$ or $-1.3$). The asymmetry allows gradient-based optimization 
to change the zero-outputting gate by changing the other gate. For this reason, we name these gates self-resurrecting activation units.

The asymmetry of SRAU may appear counter-intuitive. We contrast SRAU with a ``normalized'' version where the two gates $\tilde{B}^{\text{vis}}[i, j]$ and $\tilde{B}^{\text{lan}}[i, j]$ become symmetric. 
\begin{equation} 
\label{eq:srau-2}
\begin{split}
\tilde{B}^{\text{vis}}[i, j] & = \frac{B^{\text{vis}}[i, j] }{B^{\text{vis}}[i, j] + B^{\text{lan}}[i, j] } , \\  
\tilde{B}^{\text{lan}}[i, j] & = \frac{B^{\text{lan}}[i, j] }{B^{\text{vis}}[i, j] + B^{\text{lan}}[i, j] }.
\end{split}
\end{equation} These gates lose the asymmetry that enables the self-resurrecting property. 

In Figure \ref{fig:SRAU}, we visualize OCG, SRAU, and normalized SRAU. In ablation experiments, we show that SRAU outperforms than both OCG and normalized SRAU. 

\subsection{The Architecture and Training of VisualGPT}
For completeness, we introduce the overall architecture for VisualGPT. 
The image encoder comprising $K$ Transformer layers. Given an image, we extract objects in the image using an off-the-shelf object detection network. After that, we feed the spatial location into the image encoder. As such, the image encoder outputs $I$ of dimension $S \times O \times K$.

The caption decoder contains $M$ layers and its parameters are initialized from a PLM. We insert the encoder-decoder module, which is randomly initialized. We also apply meshed connections between the encoder and the decoder like in $\mathcal{M}^{2}$ Transformer. The network is trained to maximize the probability of the next token $w_t$ conditioned on tokens $w_1, \ldots, w_{t-1}$ and the encoder output $I$. After a predefined number of epochs on supervised learning, we switch to self-critical reinforcement learning \cite{self_critical} with CIDEr as the reward.




\begin{table*}[h!]
\centering 
\begin{tabular}{@{}l c| lllll |lllll   @{}} %

\multicolumn{2}{c}{}      &  \multicolumn{5}{c}{COCO} & \multicolumn{5}{c}{Conceptual}\\
Method &  PLM & B1 & B4 & M & R & C &  B1 & B4 & M & R & C \\  [0.5ex] 
\toprule
&\multicolumn{10}{c}{\textbf{0.1\% training data }}  \\

Transformer \cite{vaswani2017attention} & None & 57.4 & 13.1 & 16.7 & 40.7 & 40.8 & 12.4 & 2.4 & 4.9 & 15.2 & 21.2  \\ 
$\mathcal{M}^{2}$ Transformer \cite{cornia2020meshed}& None  & 56.9 & 13.1 & 16.9 & 40.6 & 40.9  &  13.1 & 2.8 & 4.8 & 15.5 & 23.5\\

AoA Transformer \cite{huang2019attention} & None &    56.6   & 13.5   & 15.9 &  40.7 &  38.4 & 11.4 & 2.4 & 4.6 & 14.7 & 20.9 \\
X-Transfomrer \cite{x-linear}  & None & 56.7   & 12.9  & 16.5 &  40.6 &  40.4 & 12.8 & 2.7 & 4.7 & 15.3 & 23.1 \\
OSCAR \cite{oscar} & BERT& 53.8 &11.9& 17.1& 39.5 &41.0 & 12.2 & 2.4 & 4.3 & 14.8 & 21.9 \\

\rowcolor{Gray}
Transformer  & GPT & 56.8 & 15.3 & 17.0 & 41.2 & 42.9 & 13.2 & 2.5 & 5.0 & 15.1 & 21.9 \\ 
\rowcolor{Gray}
$\mathcal{M}^{2}$ Transformer & GPT  & 54.9 & 14.7 & 16.6 & 41.1 & 41.0 & 11.9 & 2.6 & 4.9 & 15.4 & 24.0\\
\rowcolor{Gray}
AoA Transformer & GPT & 55.5 & 14.4 & 16.2 & 40.7 & 40.1  &  11.8 & 2.8 & 4.6 & 13.9 & 20.5 \\

\rowcolor{Gray}
VisualGPT (Normalized SRAU) & GPT & 55.7 & 15.0 & 16.8 & 41.2 & 42.4 & 13.3 & 2.9 & 5.1 & 15.8 &25.8 \\
\hline
VisualGPT (Our SRAU) & GPT & \textbf{58.2} & \textbf{16.4}& \textbf{18.5} & \textbf{41.9} & \textbf{45.1} & \textbf{13.9} & \textbf{3.2} &  \textbf{5.6} &  \textbf{16.7} & \textbf{27.7} \\

&\multicolumn{10}{c}{\textbf{0.5\% training data  }}  \\
Transformer & None & 62.8 & 18.8 & 19.4 & 25.2 & 59.2& 13.2 & 3.3 & 5.5 & 16.3 & 29.6\\ 
$\mathcal{M}^{2}$ Transformer  & None  & 63.3 & 19.4 & 19.8 & 45.6 & 61.3 & 14.5 & 3.6 & 6.0 & 17.1 & 32.0 \\
AoA Transformer  & None & 63.5 & 20.2  & 19.4 & 45.8 & 63.9 & 13.8 & 3.3 & 5.6 & 17.9 & 31.8 \\
X-Transformer  & None & 62.9& 19.0  & 19.6 & 45.7 & 62.0 & 14.2 & 3.5 & 5.8 & 17.3 & 32.1 \\
OSCAR & BERT &  59.2 & 18.0 & 21.0 & 45.3 & 60.2 & 14.4 & 3.7 & 6.1 & 17.2 &33.5 \\

\rowcolor{Gray}
Transformer  & GPT & 65.1 & 21.8 & 20.6 & 46.6 & 69.5 & \textbf{16.2} & 3.8 & 6.5 & 18.3 & 35.6 \\ 
\rowcolor{Gray}
$\mathcal{M}^{2}$ Transformer & GPT  & 64.7 & 21.8 & 20.7 & 47.1 & 68.5 & 13.9 & 3.6 & 6.0 & 17.2 & 34.1 \\
\rowcolor{Gray}
AoA Transformer & GPT &  64.2 & 21.2 & 20.5 & 46.5 & 67.2 & 14.8 & 3.6 & 6.2 & 17.6 & 34.1 \\
\rowcolor{Gray}

\rowcolor{Gray}
VisualGPT (Normalized SRAU) & GPT & 65.3 & 21.8 & 20.9 & 47.0 & 69.3 & 14.9 & 3.9 & 6.1 & 18.0 & 35.9 \\
\hline
VisualGPT (Our SRAU) & GPT & \textbf{66.2} & \textbf{22.1} & \textbf{21.1} & \textbf{47.3}& \textbf{70.3} & 15.9 & \textbf{4.2} & \textbf{6.7} & \textbf{18.5} & \textbf{37.2} \\

&\multicolumn{10}{c}{\textbf{1\% training data  }}  \\

Transformer  & None & 66.0 & 21.9 & 21.1 & 47.3 & 71.9 & 13.9 & 3.7 & 6.3 & 18.1 & 37.9\\ 
$\mathcal{M}^{2}$ Transformer & None  & 67.1 & 23.4 & 21.3 & 48.3 & 73.0 & 16.0 & 4.1 & 6.8 & 18.9 & 39.8 \\
AoA Transformer  & None & 67.6 &  23.6  & 21.5 & 48.4 & 75.5 & 14.9 & 4.1 & 6.5 & 18.6 & 39.0\\
X-Transformer  & None & 67.0 &  23.6  & 21.2 & 48.1 & 47.1 & 15.6 & 4.0 & 6.6 & 18.7 & 39.5\\

OSCAR  & BERT  & 67.2 & 23.3 & 22.5 & 49.1 &78.4 & 16.1 & 4.2 & 6.7 & 18.9 &40.6  \\

\rowcolor{Gray}
Transformer & GPT & 68.5 & 25.1 & 22.1 & 49.0 & 80.5 & \textbf{17.8} & 4.2 & 6.7 & 19.0 & 40.2\\
\rowcolor{Gray}
$\mathcal{M}^{2}$ Transformer & GPT  & 68.2 & 25.0 & 22.4 & 49.2 & 80.4 & 15.4 & 3.9& 6.5 & 17.9 & 39.1 \\
\rowcolor{Gray}
AoA Transformer  & GPT  & 68.5 & 24.6 & 22.0 & 48.6 &78.4 & 15.4 & 3.9 & 6.5 & 17.9 & 38.5 \\
\rowcolor{Gray}
VisualGPT (Normalized SRAU) & GPT & 68.7 & 25.2 & 22.3 & 49.2 & 80.6& 15.3 & 4.2 & 6.7 & 18.3 & 40.3 \\
\hline
VisualGPT (Our SRAU) & GPT & \textbf{69.5} & \textbf{25.6} & \textbf{22.6} & \textbf{49.6} & \textbf{80.9} & 16.3 & \textbf{4.3} & \textbf{6.9} & \textbf{19.3} & \textbf{40.9} \\ 

\vspace{0in}
\end{tabular}
\caption{
Performance of the compared methods training on $0.1\%$, $0.5\%$ and $1\%$ of MS COCO and Conceptual Caption image-caption pairs. The best performance in each configuration is in bold. Ablated models are marked in \colorbox{Gray}{gray}.}
\label{tab:data_efficiency}
\end{table*}

\section{Experiments}
\subsection{Datasets and Evaluation Metrics}

We evaluate our model on three datasets, MS COCO  \cite{lin2014microsoft}, Conceptual Captions \cite{sharma2018conceptual}, and IU X-ray \cite{iuxrayDataset}. MS COCO contains $123,287$ images and each of them is annotated with $5$ different captions. We follow the Karpathy split \cite{karpathy2015deep} for the validation and test set. 
The Conceptual Captions dataset \cite{sharma2018conceptual} contains 
around 3.3M images for training and 28K for validation, with much higher diversity than COCO. As the test data is not publicly available, we instead use the public validation data as our test set, and randomly sample $5000$ different image-caption pairs from the training set as the validation set.
To create the small training data setup for MS COCO and Conceptual Captions, we randomly sample $0.1\%$, $0.5\%$ and $1\%$ image-caption pairs as training data, which matches to (567, 2,835 and 5,670 pairs) for COCO and (3,300, 16,500, and 33,000 pairs) for Conceptual Captions. We repeat the experiments 4 times with different random seeds, and report the average performance. 
We report metrics for BLEU \cite{papineni2002bleu}, METEOR \cite{banerjee2005meteor}, ROUGE\cite{lin2004rouge}, and CIDEr \cite{vedantam2015cider}.

IU X-ray \cite{iuxrayDataset} is a radiography dataset containing $7,470$ chest X-ray images and $3,955$ human-written reports. As the dataset is already small, we follow the original split, which has a training set of $5,226$ images and $2,770$ reports. Most reports have two images corresponding to the frontal and lateral viewpoints.

\subsection{Experimental Settings}

\noindent \textbf{Baselines.}  We compare our model with several state-of-the-art transformer-based models, including:
\begin{itemize}
\item Plain Transformer \cite{vaswani2017attention}. 
 \item AoA Transformer, which inserts an attention-on-attention (AoA) module \cite{huang2019attention} into every transformer layer, as depicted by Figure \ref{fig:main_model} (b). Following \cite{cornia2020meshed}, we slightly update the original AoA network in \cite{huang2019attention} by replacing the LSTM with Transformers in order to create a fair Transformer-to-Transformer comparison. 
\item  $\mathcal{M}^{2}$ Transformer \cite{cornia2020meshed}, which proposes a meshed connection between encoder and decoder and is one of the best-performing models on MS COCO.

\item  X-Transformer \cite{x-linear}, which employs bilinear pooling to selectively capitalize on visual information and is one of best-performing models on MS COCO.

\item OSCAR \cite{oscar}, which finetunes BERT initialization on image-language dataset.
\end{itemize}
Since VisualGPT has GPT as the pretrained decoder, for fair comparisons, we also create variants of Transformer, AoA Transformer and $\mathcal{M}^2$ Transformer with GPT as the decoder.
For VisualGPT, we set  $\tau$ to 0.2 in \emph{all} experiments. We also explored the effect of different $\tau$ and find $\tau$ in the range of $[0, 0.2]$ to offer the right level of sparsity.
For all other baselines, we tune the hyperparameters on the validation set of MS COCO. We train our model and all the baselines in reinforcement learning setting following the work in \cite{cornia2020meshed}. Please see the supplemental material for more details on hyperparameters and experimental results.  


\begin{table}[t!]
\centering
\small
\setlength{\tabcolsep}{3pt}
\begin{tabular}{@{}lccccccc@{}}
Models & B-1 &  B-2 &  B-3 &  B-4 & R& M& C \\
\toprule
Att2in  & 22.4 & 12.9 & 8.9 & 6.8 & 30.8 & - & 29.7\\
CoAtt  & 45.5 & 28.8 & 20.5 & 15.4 & 36.9 & - & 27.7\\
HRGR  & 43.8 & 29.8 & 20.8 & 15.1 & 32.2 & - & 34.3\\
CMAS-RL  & 46.4 & 30.1 & 21.0 & 15.4 & 37.1 & - & 27.5 \\
Chen \textit{et al.}  & 47.0 & 30.4 & 21.9 & \textbf{16.5} &37.1 &  18.7 &-\\ 
\midrule
VisualGPT (ours)  & \textbf{48.0} &\textbf{ 31.3} &   \textbf{22.2} & 15.9 &  \textbf{37.4}  &  \textbf{20.5}  & \textbf{49.7} \\
\end{tabular}
\label{MedicalData}
\caption{Performance on the IU X-ray dataset.}
\end{table}

\subsection{Quantitative Results}


\noindent \textbf{Small In-domain Training Data.}  Results on MS COCO and Conceptual Captions are presented in Tables \ref{tab:data_efficiency}. VisualGPT outperforms the best-performing baseline model by $4.1$ CIDEr when trained on $0.1\%$ of MS COCO data, $6.4$ CIDEr when trained on $0.5\%$ data and $2.5$ CIDEr with $1\%$ training data. On Conceptual Caption dataset, VisualGPT also outperforms all the baselines. It outperforms the best baseline model by $4.2$ CIDEr under $0.1\%$ training data, $3.5$ CIDEr under $0.5\%$ data and $0.3$ CIDEr under $1\%$ data.

\vspace{0.1in}
\noindent \textbf{Comparison with BERT-based model.} We compared with OSCAR \cite{oscar} which is a BERT-based \cite{devlin2018bert} model with good performing results in many benchmarks. We run their model without pretraining on a large-scale image-language corpus for the fair comparison with our model. The main difference between BERT and GPT is their different pretraining objectives, where BERT uses masked language modeling and GPT is the autoregressive prediction of the next word. GPT has more similar learning behaviors to the image captioning model compared to BERT since they are both optimized by autoregressively generating the next language word. The experimental result in Table \ref{tab:data_efficiency} shows that VisualGPT is better than OSCAR in both datasets, which confirms our selection choice of using GPT as a decoder.

\begin{table}[t]
\small
\setlength{\tabcolsep}{3pt}
\centering 
\begin{tabular}{@{}l c c c c c@{}} 
Models &  B-1 & B-4 & M &R & C  \\  [0.5ex] 
\toprule
Kim \textit{et al.}  \cite{kim-etal-2019-image} & 58.1 &  13.4 &  15.9 &-& 36.0 \\
Kim \textit{et al.} + unpaired & 63.0 & 18.7& 20.7&-&55.2 \\
\midrule
\rowcolor{Gray2}
Gu \textit{et al.} \cite{gu2018unpaired}& 46.2&5.4&13.2&-&17.7 \\
\rowcolor{Gray2}
Feng \textit{et al.} \cite{feng2019unsupervised} & 58.9 &18.6&17.9&-&54.9 \\
\midrule
VisualGPT (ours) & \textbf{67.1} & \textbf{24.3}& \textbf{21.9} & \textbf{48.6} & \textbf{75.8} \\
\end{tabular}
\caption{Comparison with unsupervised and semi-supervised learning methods using Kim \textit{et al.}'s split of MS COCO. Kim \textit{et al.} employ only $1\%$ images for training in contrast to $1\%$ image-caption pairs from Table \ref{tab:data_efficiency}. Note that Kim \textit{et al.} + unpaired also use the rest of training data as unpaired images and texts. The \colorbox{Gray2}{gray} shading denotes baselines that use a large amount of unpaired images and texts during training.} 
\label{Kim} 
\end{table}

\vspace{0.1in}
\noindent\textbf{Medical Report Generation.} We compared VisualGPT against state-of-the-art medical report generation models including Att2in \cite{self_critical}, CoAtt \cite{jing2018automatic}, HRGR \cite{li2018hybrid}, CMAS-RL \cite{chestXrayReport} and the model from Chen \textit{et al.} \cite{chen2020generating}. This dataset only contains around $2,770$ medical reports in the training set, which is less than $1\%$ COCO data and poses a data-efficiency challenge. We follow the same experimental setting as in \cite{chen2020generating}. The results show that VisualGPT outperforms the baselines for most evaluation metrics and creates a new state-of-the-art. It shows the value of leveraging GPT knowledge into the highly specific domain which has very ``expensive" and insufficient paired data. We hope our finding could inspire future work in other domains.

\vspace{0.1in}
\noindent\textbf{Comparison Against Semi-supervised and Unsupervised Methods.} 
Kim \textit{et al.} \cite{kim-etal-2019-image} proposed a semi-supervised learning method to improve the data efficiency of image captioning. They used $1\%$ of images  and all their captions as training data, rather than $1\%$ of all the image-caption pairs in Table \ref{tab:data_efficiency}, hence they cover less images since each image is associated to more than 1 caption. For Kim \textit{et al.} + unpaired, they also employ the other $99\%$ of MS COCO as unpaired images and captions for training. We replicate their setup by only training with 1\% of images. As shown in Table \ref{Kim}, without using additional unpaired images and captions, the proposed VisualGPT method outperforms Kim \textit{et al.} \cite{kim-etal-2019-image} by $20.6$ CIDEr score. 

We also compare VisualGPT against unsupervised methods of Gu \textit{et al.} \cite{gu2018unpaired} and Feng \textit{et al.} \cite{feng2019unsupervised}, which use tens of millions of unpaired images and captions. Even though these are not fair comparisons, it is encouraging to see VisualGPT surpassing these baselines by utilizing the supervision of only $1133$ training images.

\begin{figure}
\centering
 \includegraphics[width=\linewidth]
                  {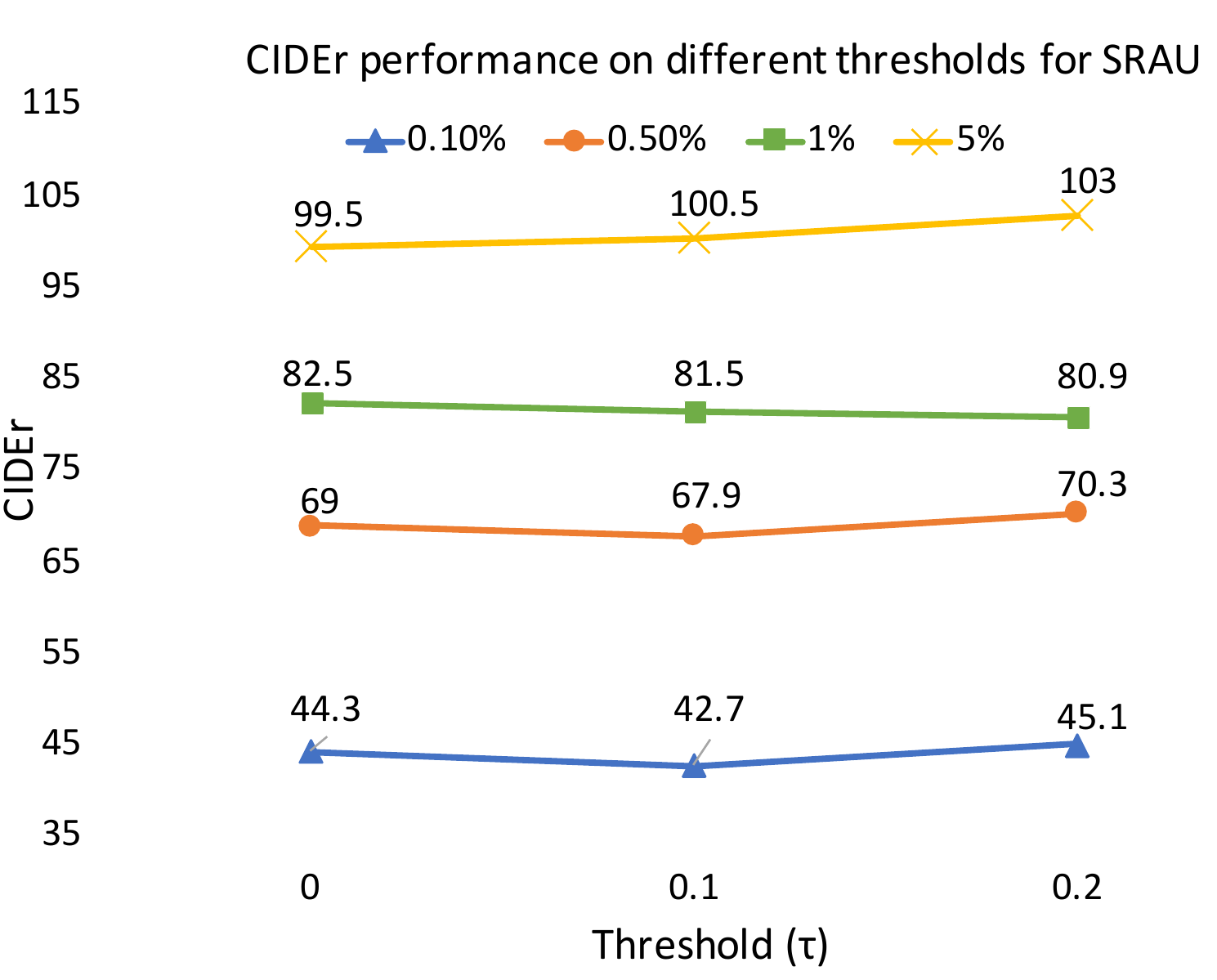}
\caption{CIDEr performance v.s. different thresholds $\tau$ with 0.1\% 0.5\%, 1\% and 5\% training data.}
\label{fig:cider_threshold}
\end{figure}

\subsection{Ablation Studies}
\label{sec:ablation_study}

\noindent \textbf{Ablation on cross-attention:} To fairly compare our SRAU with other cross-attention mechanisms in the baselines, we also initialize their decoder with 12-layer GPT and keep the same encoder as VisualGPT. We contrast between plain cross-attention, meshed cross-attention, and attention-on-attention (AoA) modules. For AoA Transformer, we add the AoA module on top of cross-attention.   Table \ref{tab:data_efficiency} shows the results, which demonstrate that SRAU is better than other cross-attention modules in exploiting the GPT knowledge within the image-caption task. 




\vspace{0.1in}
\noindent\textbf{Ablation on SRAU:} We create an ablation called \emph{Normalized SRAU}, where we replace the SRAU with the normalized SRAU (see Figure \ref{fig:SRAU}) and use GPT2 initialization.  We provided the results in table \ref{tab:data_efficiency}. The normalized SRAU results in substantially lowered performance, decreasing CIDEr from full VisualGPT by $2.7$, $1.0$, and $0.3$ respectively on the three setups on MS COCO, and it also decreases from Full VisualGPT by $2.2$, $1.3$ and $0.6$ respectively on Conceptual Caption. This demonstrates that the self-resurrecting property is beneficial for learning from small data. We experimented with Leaky ReLU and GELU, which ameliorate zero gradients, but the training crashed due to the lack of upper limits for function values. 


We explored different $\tau$ among (0, 0.1 0.2) and show their CIDEr performance on different percentage of COCO training data in the Figure \ref{fig:cider_threshold}. $\tau$=0 is equivalent to ordinary complementary sigmoid gates. We can observe that $\tau$ = 0.2 can give us the best performance in most cases, indicating the usefulness of incorporating sparsity in our SRAU complementary gates.


\begin{table}[t]
\small
\setlength{\tabcolsep}{3pt}
\centering
\begin{tabular}{@{}c c c c@{}}
Method & 0.1\% data & 0.5\% data &    1\% data \\ 
\toprule
Transformer  & 18.4\% & 17.2\% &    16.8\% \\
AoA Transformer  & 11.5\% & 20.9\% &       25.0\% \\
$\mathcal{M}^2$ Transformer  & 30.9\%& 22.8\%   &    20.8\% \\  
\midrule
VisualGPT & \textbf{39.2\%} & \textbf{39.1\%}  &     \textbf{37.4\%}  \\
\end{tabular}
\caption{The percentage of votes received by VisualGPT and baseline models under different quantity of training data.}
\label{aws}
\end{table}

\begin{table}[t]
\small
\setlength{\tabcolsep}{3pt}
\centering 
\begin{tabular}{@{}c c c c c c c @{}} 
\multicolumn{6}{c}{\textbf{Q1. Does the caption miss things shown in the image?}} \\
Answer & Ours & $\mathcal{M}^{2}$Transformer & Transformer & AoA & GT  \\   
\toprule
No & 719 & 624 & 633&   621& 973 \\
Yes & 367 & 438 & 456 &447&  73  \\  \hline
No Rate & \textbf{0.66} & 0.59 & 0.58&0.58 & 0.93 \\
\specialrule{0.5pt}{0.1pc}{0.8pc}
\multicolumn{6}{c}{\textbf{Q2. Does the caption describe things not in the image?}} \\
Answer & Ours& $\mathcal{M}^{2}$Transformer & Transformer & AoA & GT  \\   
\toprule
No & 720 &692 &633 & 655 & 448 \\
Yes & 360 & 418 & 423 &  412 & 43 \\  \hline
No Rate & \textbf{0.67} &0.62 & 0.60 & 0.61 & 0.96 \\
\end{tabular}
\caption{Human evaluation of object hallucination and omission. GT denotes the ground-truth captions.}
\label{q2} 
\end{table}


\begin{figure}[t]
\centering
\scalebox{0.95}{
 \includegraphics[width=\linewidth]{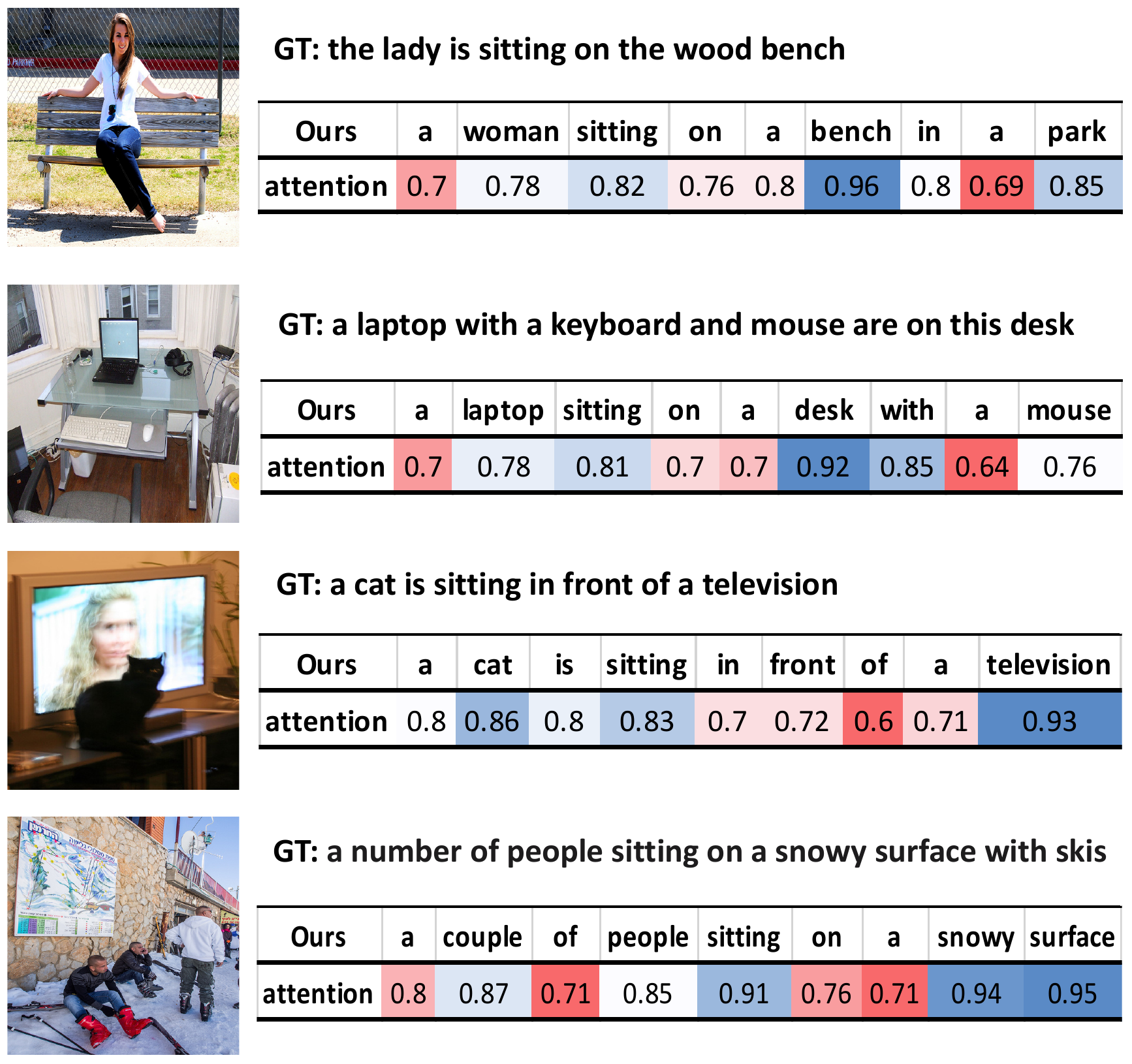}
}
\caption{Visual scores of words in generated captions. We show the raw visual scores and highlight them according to normalized visual scores. High visual scores are in blue and low scores in red.  }

\label{fig:balaned_attention_analysis}
\end{figure}

\subsection{Human Study}
In addition to automatic evaluation metrics, we conduct two human studies to further evaluate the quality of generated captions. In the first study, we asked participants directly for preference over generated captions. We randomly selected 250 test images from the three setups of $0.1\%$, $0.5\%$, and $1\%$ training data. For every image, we generated one caption from VisualGPT and each of three high-performing baselines from Table \ref{tab:data_efficiency}, Transformer \cite{vaswani2017attention}, $\mathcal{M}^{2}$ Transformer \cite{cornia2020meshed}, and AoA Transformer \cite{huang2019attention}, all with three decoder layers. Every image was evaluated by 5 different Turkers, who chose the caption that most accurately described the image content.  We received $3750$ ($250$ images $\times \, 5$ Turkers $\times \, 3$ setups) valid responses. 

 We summarize the results in Table \ref{aws}.
Overall, the captions generated by VisualGPT received the largest share of votes, $39.2 \%$ for the $0.1\%$  training data split, $39.1\%$ for the $0.5\%$ split, and $37.4\%$ for the $1\%$ split. For each training setup, we conducted Pearson’s Chi‑square test~\cite{pearson1900x}, which shows the differences are statistical significant with $p < 0.05$ in all cases. 

In the second study, we evaluate if using pretrained language models introduces excessive linguistic prior that could cause the known object hallucination problem \cite{rohrbach2018:object_hallucination}. 
From the models trained using $1\%$ COCO data. We randomly sampled $250$ images with the generated caption from each model. For each image, we asked $5$ different participants if the caption (1) described non-existent objects or (2) missed objects existing in the image. To catch random clickers, we created $5$ images with verified captions, so that we knew the right answers of these questions. Participants who answered these questions wrongly were considered unreliable and removed from the results.

The results are in Table \ref{q2}. Compared to the baselines, VisualGPT has less hallucination and higher coverage of objects. The study also finds that the ground-truth captions has the least amount of hallucination and highest coverage of objects in the image. This finding lends positive support to the validity of the experimental protocol.  

\begin{figure}
\centering
 \includegraphics[width=\linewidth]
                  {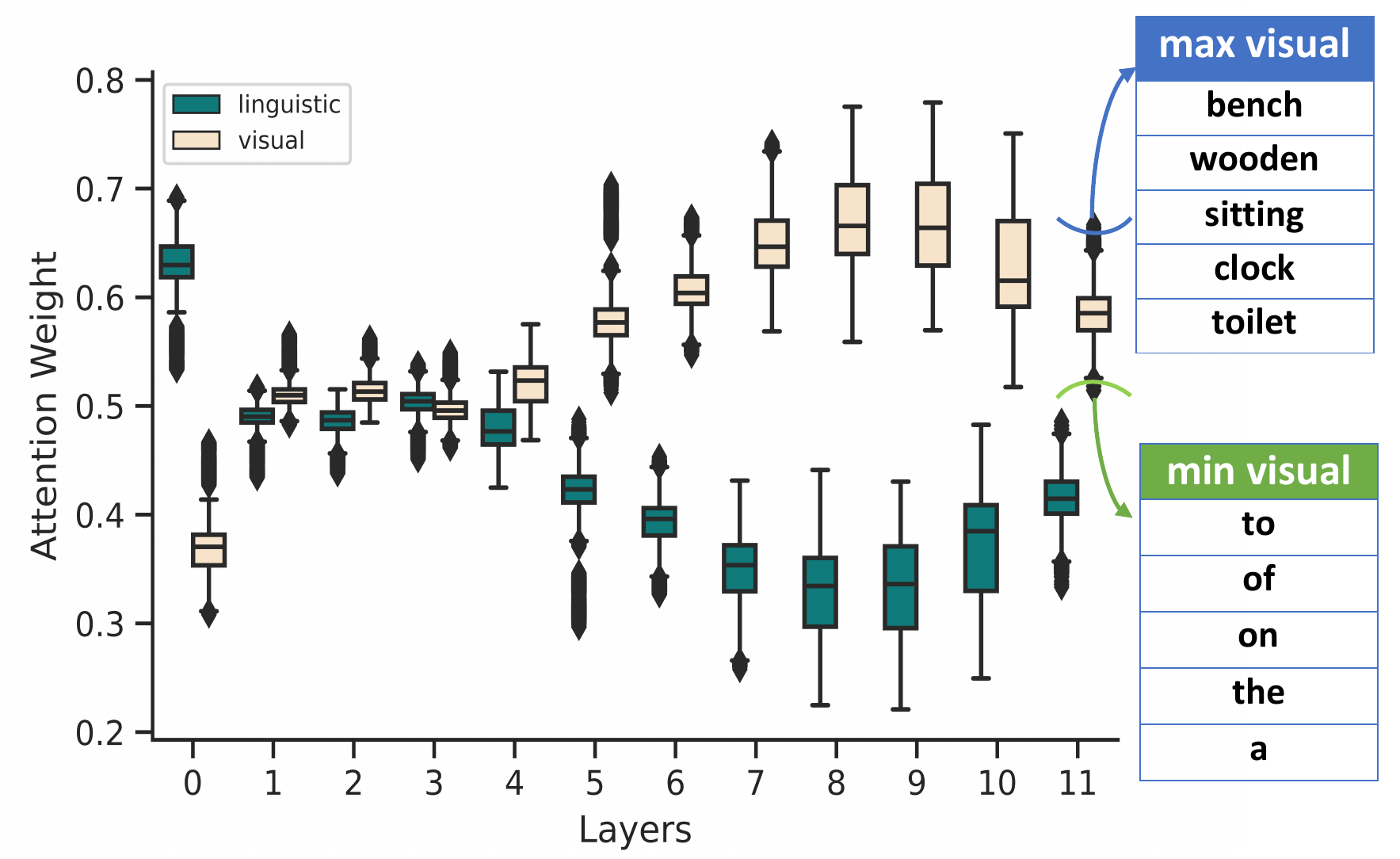}
\caption{Distributions of linguistic attention ($B^{\text{lan}}$) and visual attention ($B^{\text{vis}}$) at every decoding layer. We also show the words generated with the highest and lowest visual attention.}
\label{fig:09p_visual_average}
\end{figure}

\subsection{Analysis}
In this section, we visually examine examples from the VisualGPT model trained on 1\% of MS COCO. First, we show example captions generated by VisualGPT in Figure \ref{fig:balaned_attention_analysis} and the associated $B^{vis}$ at the last decoder layer. Note that for every word generated, we have a $768$-dimensional visual gate vector, which is a slice of $B^{vis}$ at different decoding time steps. We take the mean of the gate vector as the visual score for that word. After that, we normalize the visual scores across the dataset to the $[0,1]$ interval and highlight the words accordingly. Blue indicates high visual scores and red indicates low visual scores. We observe that,  in agreement with our intuition, VisualGPT assigns high visual scores to words like ``desk'' and ``snowy surface'' and low visual scores to determiners and prepositions.

In Figure \ref{fig:09p_visual_average}, we plot the distribution of $B^{vis}$ and $B^{lan}$ at every decoder layer as a box-and-whisker diagram. We also show the words with the highest and lowest visual scores, which are again in line with our expectations. Additionally, we observe that, going from layer 0 to layer 9, the decoder  makes increasing use of visual information, but the uppermost layers, 10 and 11, make more balanced use of information. We hypothesize that the low layers focus on low-level linguistics like syntax, whereas the middle layers learn to fuse linguistic information with visual information. Finally, the two information sources become balanced in the uppermost layers.

\subsection{Limitation}

One limitation of our proposal is that, as experiments in the supplementary material show, the gap between baseline models and VisualGPT gradually vanishes as in-domain training data increase. The phenomenon is  more pronounced in COCO than Conceptual Captions, which has a more diverse vocabulary. We hypothesize that linguistic knowledge from pretrained models is the most useful when the training data are small and do not provide sufficient coverage of the vocabulary. 






\section{Conclusions}
 We present VisualGPT, a data efficient image captioning model which leverages the linguistic knowledge from the pretrained language model. To bridge the semantic gap between different modalities, we design a novel encoder-decoder attention mechanism with an unsaturated rectified gating function. We evaluate our model on $0.1\%$, $0.5\%$ and $1.0\%$ of MS COCO and Conceptual Captions, and IU X-ray, a small medical imaging report dataset. VisualGPT achieves the state-of-the-art result on IU X-ray and outperforms  strong baseline models. 
 
 VisualGPT may solve the realistic need when training captioning models on low-resource languages or highly specialized domains, where it could be challenging to find annotators to collect a large amount of data. 
 

\vspace{0.1in}
\noindent \textbf{Acknowledgments.}
This work is funded by  KAUST BAS/1/1685-01-0, KAUST-FCC/1/2533-17-01, and National Research Foundation Fellowship (NRF-NRFF13-2021-0006), Singapore. 


\newpage
{\small
\bibliographystyle{ieee_fullname}
\bibliography{egbib-simple,egbib-extra-simple}
}

\clearpage

\appendix

\section{Supplementary material}

\subsection{Additional implementation details}
\noindent \textbf{Image and Word Features.} Following \cite{anderson2018bottom}, we use a Faster R-CNN networks \cite{ren2015faster} with ResNet-101 \cite{he2016deep} as a backbone to train on Visual Genome dataset \cite{krishna2017visual}, and we extract a $2048$-dimensional feature vector for each object.

We use the Byte Pair Encoding (BPE) \cite{BPE}, which effectively incorporate sub-word information and is beneficial for dealing with out-of-vocabulary words. We employ learnable positional encoding and initialize token embedding from pretrained weights of GPT-2.

\noindent \textbf{Architecture and Hyperparameters.} We have $3$ layers in the encoder and $12$ layers in the decoder with $12$ heads in each layer. The hidden size $D$ in each layer is $768$. We load the GPT-2 (small) pretrained weights, which has 117M parameters into the decoder. We use the learning rate of $1e^{-4}$ under XE loss and $1e^{-5}$ during the reinforcement learning.  We train the models with the AdamW optimizer \cite{loshchilov2018fixing} and a batch size $25$. The beam size is equal to $5$. The threshold $\tau$ is tuned on the validation set for different training data.


\noindent \textbf{Training Details.} We train all the models in two steps. We first train the models with cross-entropy (XE) loss and then finetune them using reinforcement learning.  
The cross-entropy loss $\mathcal{L}_{XE}$ is the traditional autoregressive classification loss
\begin{equation} 
   \mathcal{L}_{XE} = - \sum_{t=1}^{T} log((w_{t} | w_{1:t-1})) \end{equation}
where $w_{1:T}$ represents the target ground truth sequence.

For reinforcement learning, we employ a variant of Self-Critical Sequence training \cite{rennie2017self}. Following \cite{cornia2020meshed}, we sample $L$ sentences, $\hat{w}_{1:T}^1, \ldots, \hat{w}_{1:T}^L$, with beam search and use the mean reward from the $L$ sentences as the baseline $b$. The gradient is
\begin{equation} 
   \nabla_{\theta}\mathcal{L}_{RL}(\theta) = - \frac{1}{k} \sum_{i=1}^{L} \bigg((r(\hat{w}^i_{1:T}) - b ) \nabla_{\theta} \log p(\hat{w}^i_{1:T}) \bigg)    \end{equation}
where $r(\cdot)$ represents the CIDEr-D reward. 

\begin{figure}[h]
\begin{center}
  \includegraphics[width=0.9\linewidth]{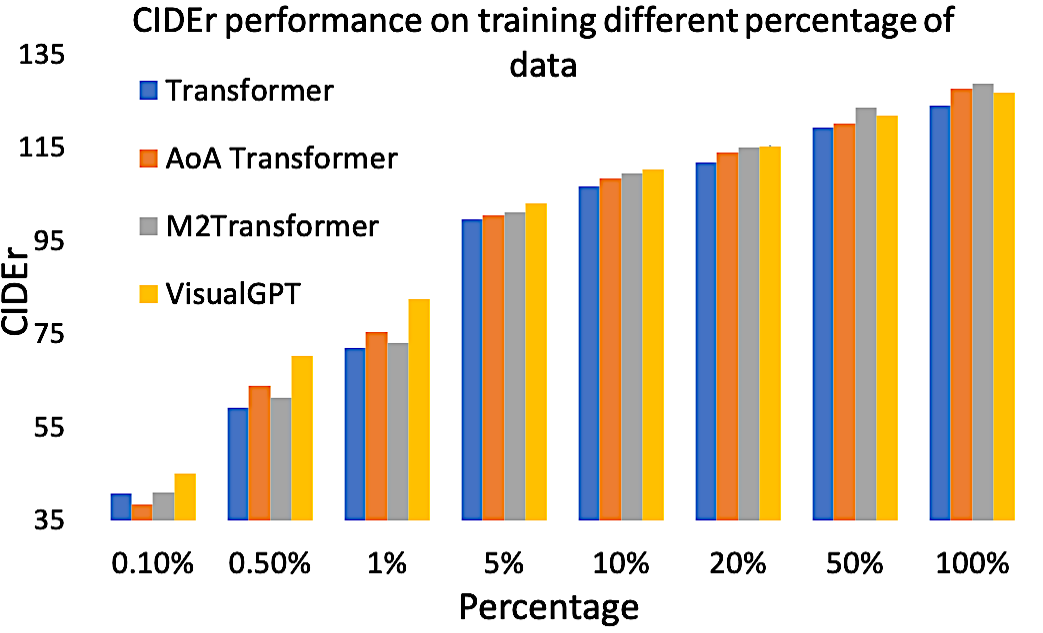}
\end{center}
\caption{Evaluation on different percentage of COCO data}
\label{fig:scale}

\end{figure}

\begin{figure}[h]
\begin{center}
  \includegraphics[width=1.0\linewidth]{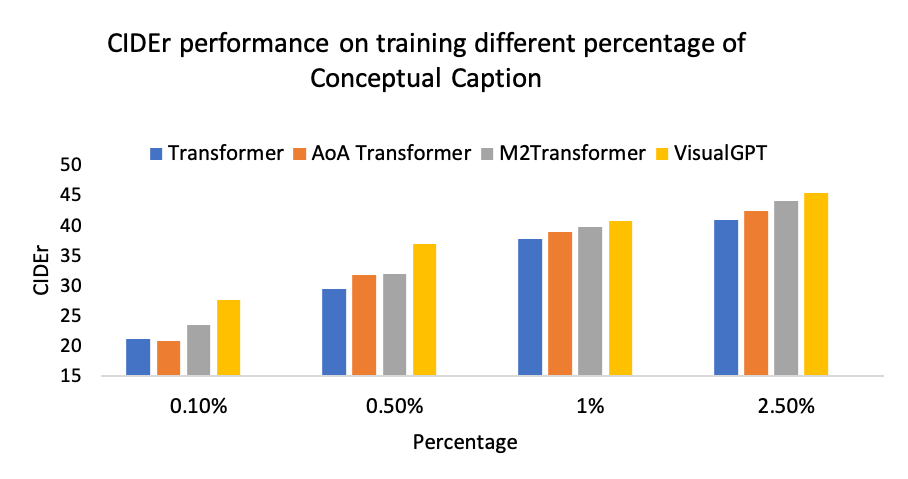}
\end{center}
\caption{Evaluation on different percentage of Conceptual Captions}
\label{fig:conceptual}

\end{figure}

\subsection{Train VisualGPT with more COCO and Conceptual Caption Datasets}
Figure \ref{fig:scale} shows other results obtained by  training networks on the $5\%$ , $10\%$, $20\%$, $50\%$ and $100\%$ (82,783 images) MS COCO data. Figure \ref{fig:conceptual} shows the performance with the data scaling up to $2.5\%$ (82,958 images) Conceptual Captions, in which the dataset scale is similar to the whole COCO data. For MS COCO, VisualGPT outperforms other baseline models when we sample $\leq 20 \%$ training data. For Conceptual Caption, VisualGPT consistently outperforms all the baselines when we sample $\leq 2.5 \%$ training images. The whole experiments  highlight our model's effectiveness on low data regimes.

On the other hand, we should also notice  that $\mathcal{M}^2$ Transformer surpasses the VisualGPT's performance when there are 50\% and 100\% COCO training data. But when we train with the same number of Conceptual images, VisualGPT continuously outperforms all the baselines. This leads us to think of the reason why VisualGPT show different performing behaviors on these two datasets. The difference between these two datasets is that the Conceptual Captions contain more diverse vocabularies and image contents. In contrast, COCO captions only cover 80 common image objects. Therefore, the appearance frequency for each word in COCO is  much higher than that in Conceptual Captions and COCO vocabulary diversity is also much lower than Conceptual Caption. We hypotheses the reason for this performance difference is that when the captions have a small coverage of each word, the caption generation will be benefited a lot from the GPT inherent knowledge and GPT can help the model quickly adapt into the new domain. But when there is a lot of in-domain data, the current image-captioning models can already generalize well on it and it potentially contradicts to the GPT original knowledge.

\subsection{Attention over Different types of words}
\begin{figure}[h]
\begin{center}
   \includegraphics[width=1.0\linewidth]{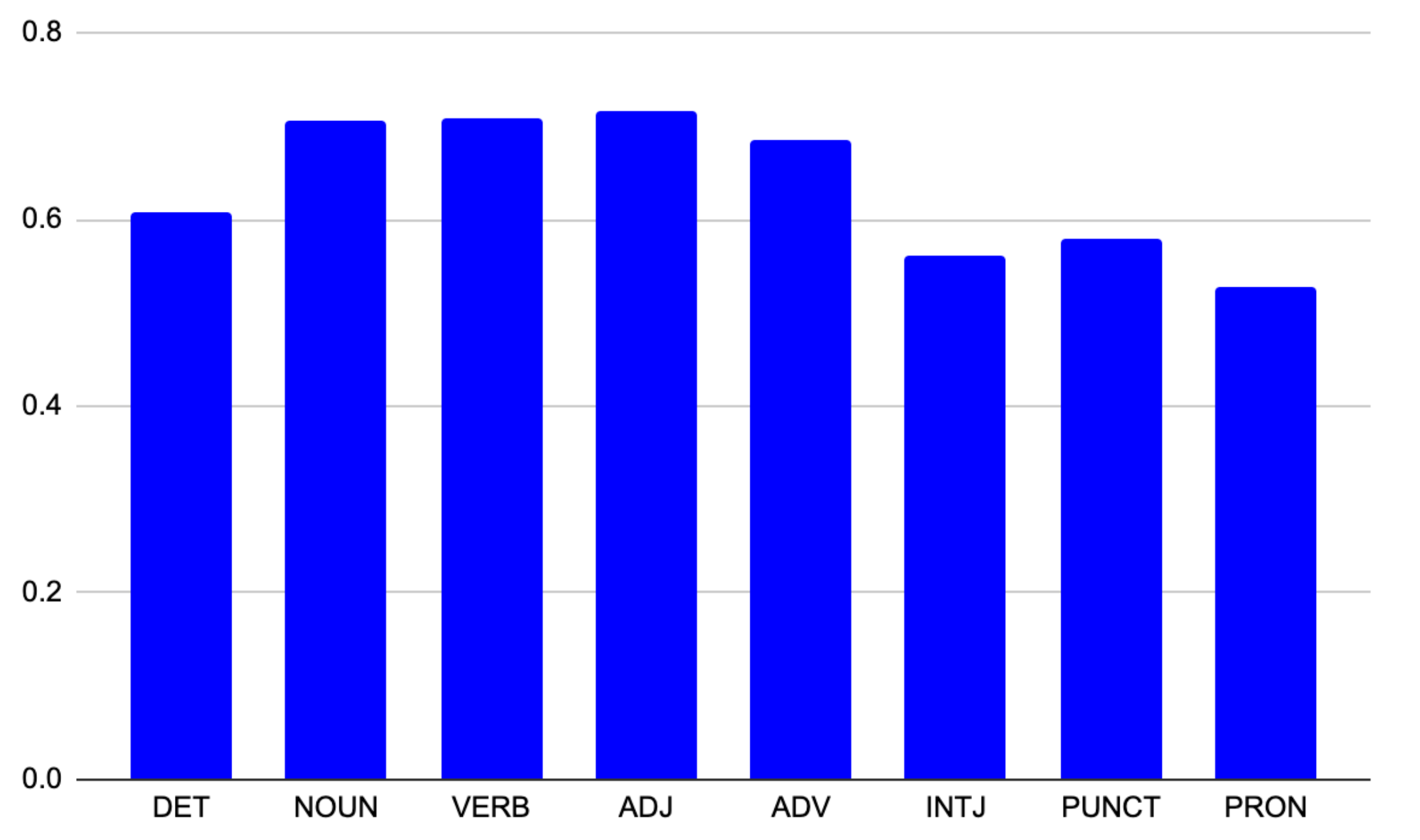}
\end{center}
\caption{Attention Scores over different part-of-speech words}
\label{fig:pos_attention}

\end{figure}

We use the Spacy parser to detect the part-of-speech of words in captions and calculate the mean value of the visual attention score. The result is presented in Fig. \ref{fig:pos_attention}. We found PoS that tend to visual content, like noun (0.71), verb (0.71) and adjective (0.72), have high visual attention scores, whereas linguistic PoS like pronoun (0.53), punctuation (0.58), and determiner (0.61) receive low attention.

\subsection{More Qualitative Examples}

\begin{figure}[t]
\centering
\begin{subfigure}{0.5\textwidth}
  \centering
  \includegraphics[width=0.87\linewidth]{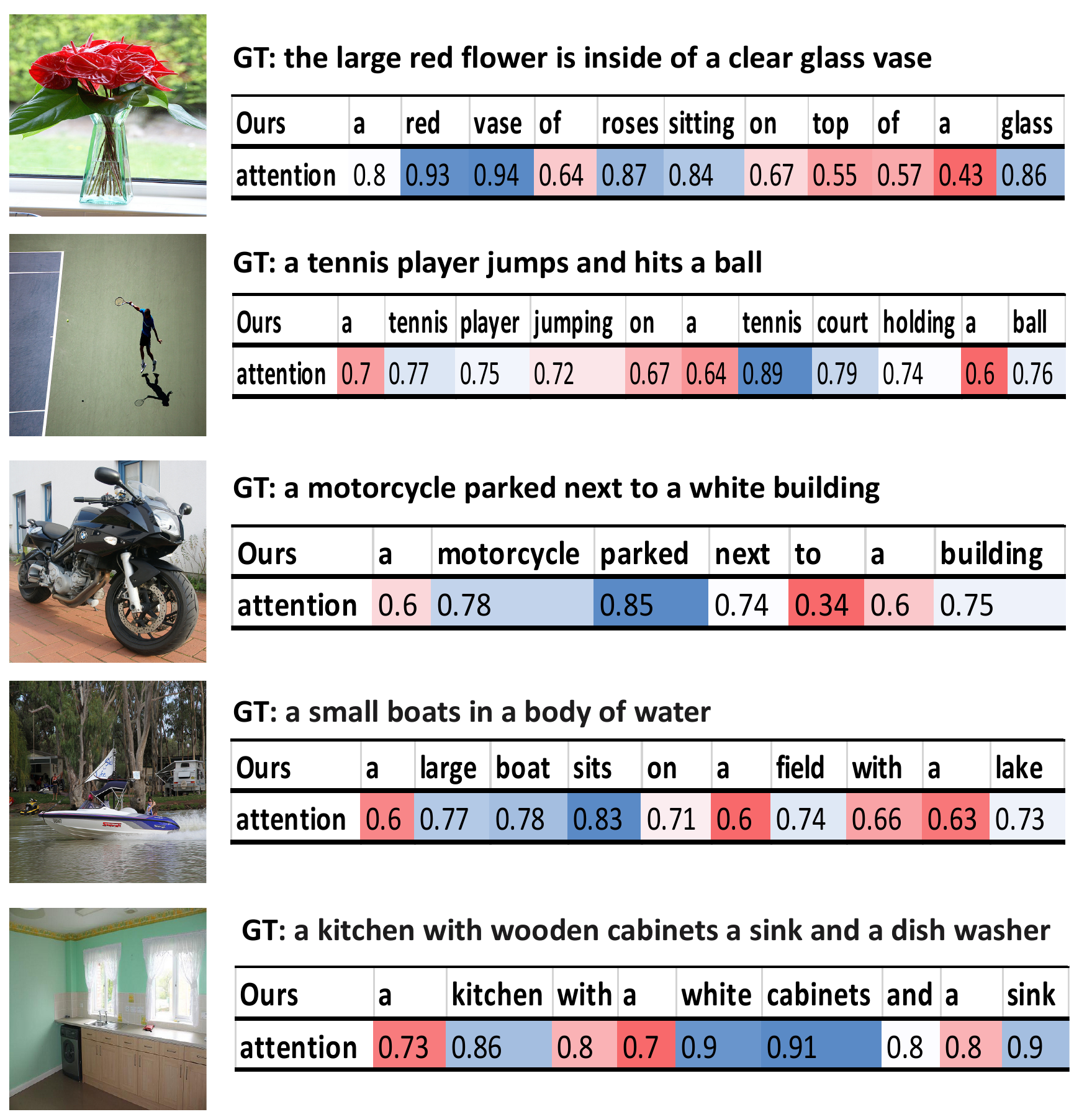}
\end{subfigure}

\begin{subfigure}{0.5\textwidth}
  \centering
  \includegraphics[width=0.87\linewidth]{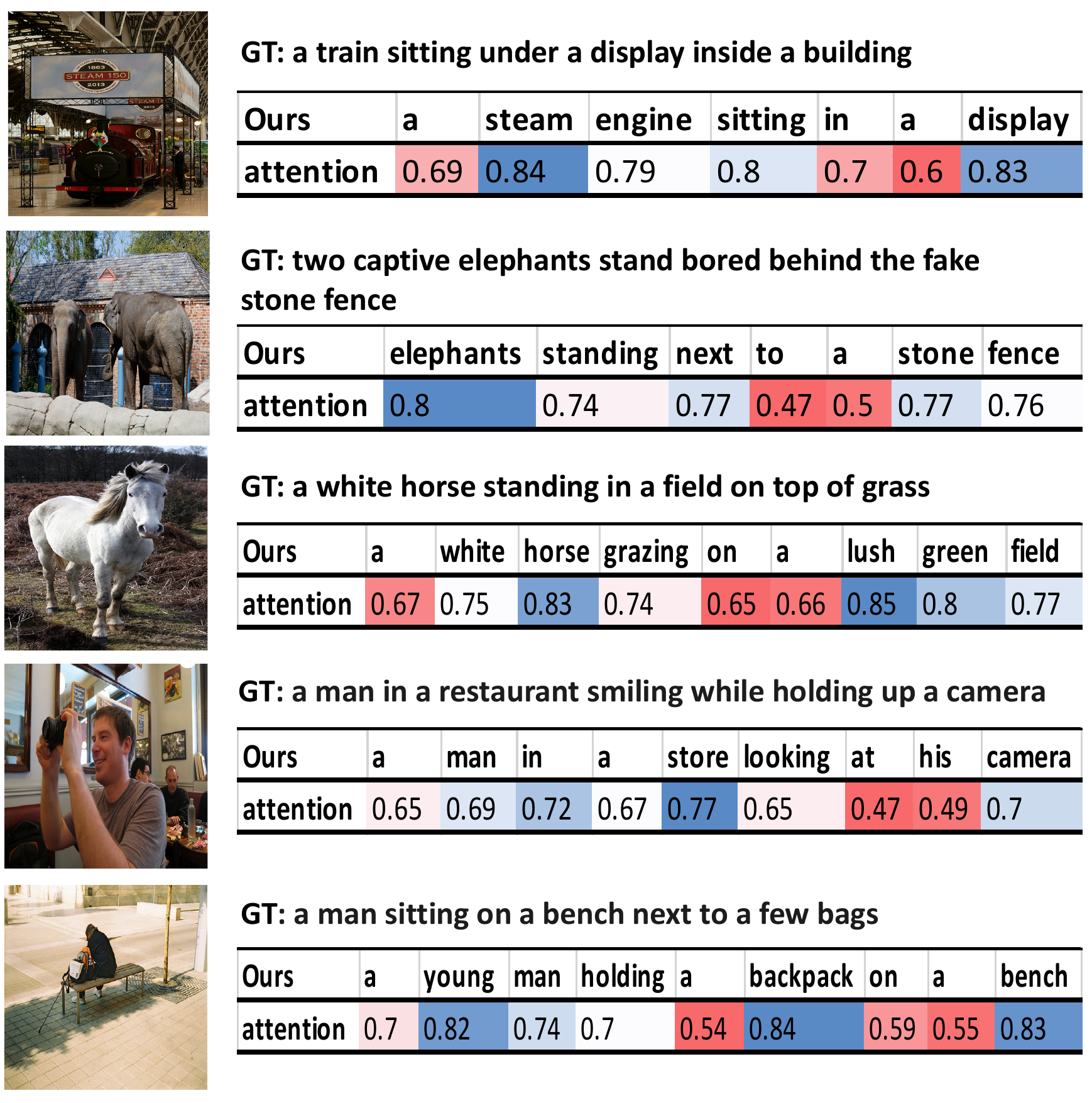}

\end{subfigure}
\caption{More examples of visual attention for each word in generated captions. High visual scores are in blue and low scores in red.}
\label{fig:attention}
\end{figure}

In Figure \ref{fig:attention}, we provide more examples of visual attentions. Blue indicates high visual scores and red indicates low visual scores. We can observe that VisualGPT assigns higher scores to words like ``steam engine'', ``elephants'', ``horse'', ``lush'' and ``cabinets'', and it assigns low visual scores to determiners and prepositions like ``to'' and ``at''.

We also show some examples of generated captions by our VisualGPT and several strong baseline models including Transformer ($3$ layers) \cite{vaswani2017attention}, $\mathcal{M}^{2}$ Transformer ($3$ layers) \cite{cornia2020meshed} and AoA Transformer \cite{huang2019attention} in the Table \ref{supp_table1}, Table \ref{supp_table2} and Table \ref{supp_table3}. Overall, we can observe that our VisualGPT is able to describe the image content more accurately than the baseline models.

\begin{table*}

\setlength{\tabcolsep}{3pt}
\begin{tabular}{lll}
\hline

Image & Generated Captions & Ground Truth \\ \hline
\begin{tabular}{l}
     \includegraphics[width=35mm,height=32mm
]{sup_image/6.jpg} 
\end{tabular}

& 
\begin{tabular}{l}
    \makecell[l]{ 
\textbf{Transformer}: a woman riding some skis on skis \\ 
\textbf{$\mathcal{M}^{2}$ Transformer}: a couple of skiers are \\standing near the snow\\
\textbf{AoA Transformer}:  a man with skis in the snow \\ 
\textbf{VisualGPT (ours)}: a group of people walk on a\\ snowy mountain} 
\end{tabular}
& 
\begin{tabular}{l}{
\makecell[l]
{\textbf{GT1}: the people are walking through snow\\ in a wooded area \\
\textbf{GT2}: two people wearing skis traveling \\through the snow\\
\textbf{GT3}: a man is walking down a path covered\\ in a snow\\
\textbf{GT4}: a couple is skiing through the \\snowy woods\\
\textbf{GT5}: a couple of people that are in a\\ snowy field} }\end{tabular}
\\
\hline

\begin{tabular}{l}
     \includegraphics[width=35mm,height=32mm
]{sup_image/7.jpg} 
\end{tabular}

& 
\begin{tabular}{l}
    \makecell[l]{ 
\textbf{Transformer}: a street that has some street \\in it\\
\textbf{$\mathcal{M}^{2}$ Transformer}: a traffic light over a street \\light under a traffic light\\
\textbf{AoA Transformer}:  a street with people on a \\city street\\
\textbf{VisualGPT (ours)}: a street with tall signs and\\ traffic signs} 
\end{tabular}
& 
\begin{tabular}{l}{
\makecell[l]
{\textbf{GT1}: a yellow traffic light above a street\\ next to houses\\
\textbf{GT2}:a street scene of an intersection with a\\ street light\\
\textbf{GT3}: a stop light hanging over an\\ intersection in a residential area\\
\textbf{GT4}: a traffic signal at an intersection is\\ suspended on wire\\
\textbf{GT5}: a street intersection with a traffic\\ light over it} }\end{tabular}
\\
\hline


\begin{tabular}{l}
     \includegraphics[width=35mm,height=32mm
]{sup_image/8.jpg} 
\end{tabular}

& 
\begin{tabular}{l}
    \makecell[l]{ 
\textbf{Transformer}: some pizza are sitting on a\\ plate\\
\textbf{$\mathcal{M}^{2}$ Transformer}: a plate with food and\\ a knife on it\\
\textbf{AoA Transformer}:  a plate of pizza on a table\\ 
\textbf{VisualGPT (ours)}: a plate of bread are served\\ on a table} 
\end{tabular}

& 
\begin{tabular}{l}{
\makecell[l]
{\textbf{GT1}: a batch of bread slices sitting on a\\ plate \\
\textbf{GT2}: a plate with some pieces of bread on it\\
\textbf{GT3}: sliced french bread is on a plat that\\ is lying on a table\\
\textbf{GT4}: bread that is sitting on a plate that is\\ on a table\\
\textbf{GT5}: a white plate with lots topped with\\ garlic bread} }\end{tabular}
\\
\hline


\begin{tabular}{l}
     \includegraphics[width=35mm,height=32mm
]{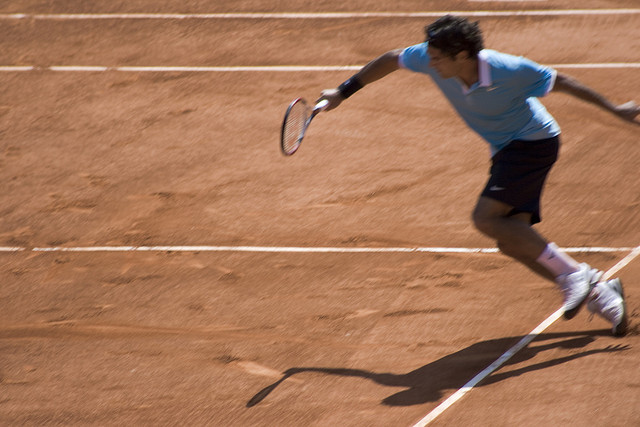} 
\end{tabular}

& 
\begin{tabular}{l}
    \makecell[l]{ 
\textbf{Transformer}: two tennis player playing tennis\\ on the ball\\
\textbf{$\mathcal{M}^{2}$ Transformer}: a tennis player about to\\ hit a ball\\
\textbf{AoA Transformer}:  a baseball players on a game\\ playing a game\\ 
\textbf{VisualGPT (ours)}: a tennis player hits a ball\\ with a racket} 
\end{tabular}

& 
\begin{tabular}{l}{
\makecell[l]
{\textbf{GT1}: a man holding a racquet on top of a\\ tennis court \\
\textbf{GT2}: a man with a tennis racket reaches\\ for a ball\\
\textbf{GT3}: a man with a tennis racket is running\\ on a court\\
\textbf{GT4}: a young man is playing a game of\\ tennis \\
\textbf{GT5}: a tennis player in a blue shirt\\ runs toward a ball}}\end{tabular}
\\
\hline


\begin{tabular}{l}
     \includegraphics[width=35mm,height=32mm
]{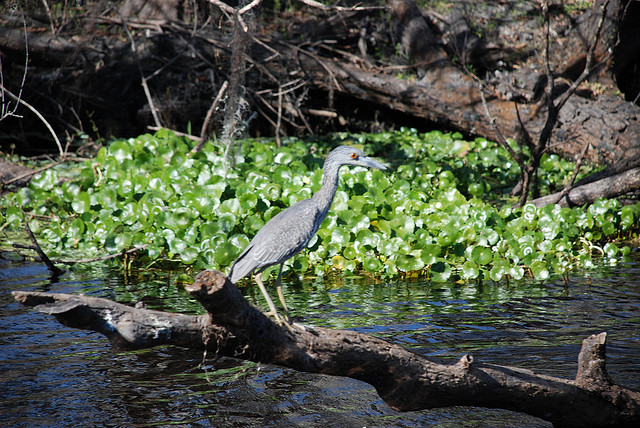} 
\end{tabular}

& 
\begin{tabular}{l}
    \makecell[l]{ 
\textbf{Transformer}: a group of birds that are\\ standing in the grass\\
\textbf{$\mathcal{M}^{2}$ Transformer}:   a flock of birds perched\\ in a tree branch\\
\textbf{AoA Transformer}:  several giraffe are standing \\next to each trees\\ 
\textbf{VisualGPT (ours)}: a bird standing in the\\ middle of a pond} 
\end{tabular}

& 
\begin{tabular}{l}{
\makecell[l]
{\textbf{GT1}: a bird is perched a top a branch over \\a river \\
\textbf{GT2}: a bird sits on a branch above a stream\\
\textbf{GT3}: a bird on top of a tree branch over\\ water\\
\textbf{GT4}: a picture of an outside region that\\ appears incredible\\
\textbf{GT5}: a bird on a fallen branch in a body of\\ water}}\end{tabular}
\\
\hline
\\

\end{tabular}
\caption{Caption generated by our VisualGPT, Transformer, $\mathcal{M}^{2}$ Transformer and AoA Transformer on $0.1\%$ MS COCO data split}
\label{supp_table1}
\end{table*}

\begin{table*}

\setlength{\tabcolsep}{3pt}
\begin{tabular}{lll}
\hline

Image & Generated Captions & Ground Truth \\ \hline
\begin{tabular}{l}
     \includegraphics[width=35mm,height=32mm
]{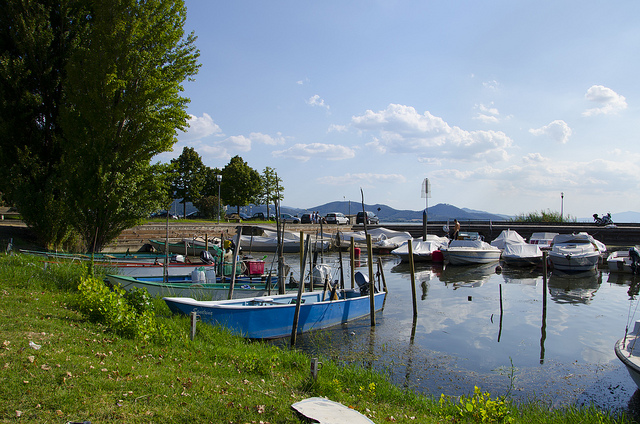} 
\end{tabular}

& 
\begin{tabular}{l}
    \makecell[l]{ 
\textbf{Transformer}: several boats are sitting in\\ the middle of a lake\\ 
\textbf{$\mathcal{M}^{2}$ Transformer}: a boat filled with \\boats floating in the water\\
\textbf{AoA Transformer}:  an empty boat that has \\ water and water \\ 
\textbf{VisualGPT (ours)}: a canal filled with boats\\ in the water} 
\end{tabular}
& 
\begin{tabular}{l}{
\makecell[l]
{\textbf{GT1}: a blue boat docked on a green lush \\shore \\
\textbf{GT2}: a small marina with boats docked there\\
\textbf{GT3}: a group of boats sitting together with\\ no one around\\
\textbf{GT4}: some boats parked in the water at\\ a dock\\
\textbf{GT5}: boats sitting around the side of a\\ lake by a tree} }\end{tabular}
\\
\\
\hline

\begin{tabular}{l}
     \includegraphics[width=35mm,height=32mm
]{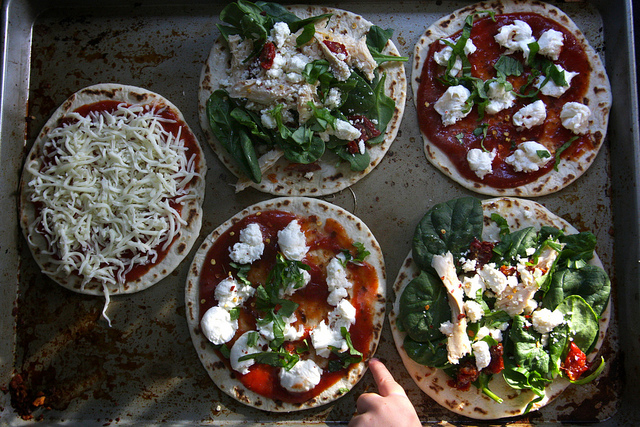} 
\end{tabular}

& 
\begin{tabular}{l}
    \makecell[l]{ 
\textbf{Transformer}:pizza slices and pizza in a \\plate covered pizza\\
\textbf{$\mathcal{M}^{2}$ Transformer}: people sitting at a\\ table eating pizza and other salad\\
\textbf{AoA Transformer}:  two pizza eating a table with \\pizza on the table \\ 
\textbf{VisualGPT (ours)}: a group of pizza on a \\iron plate with toppings} 
\end{tabular}
& 
\begin{tabular}{l}{
\makecell[l]
{\textbf{GT1}: a set of five pizzas sitting next \\to each other each with different toppings\\
\textbf{GT2}:a handful of prepared pizzas sit next\\ to each other\\
\textbf{GT3}: five uncooked pizzas with a variety\\ of different toppings\\
\textbf{GT4}: five unbaked pizzas that include \\various types of cheeses\\
\textbf{GT5}: five different pizzas are being \\prepared over a metal tray} }\end{tabular}
\\
\hline


\begin{tabular}{l}
     \includegraphics[width=35mm,height=32mm
]{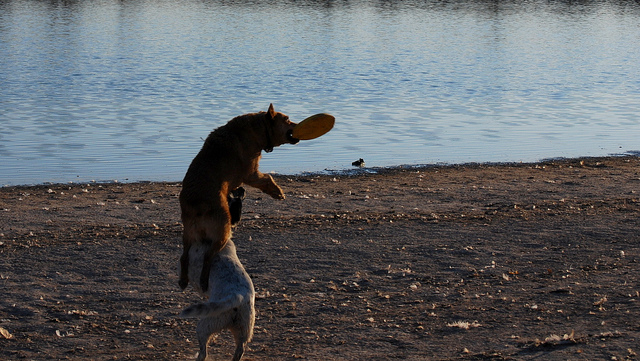} 
\end{tabular}

& 
\begin{tabular}{l}
    \makecell[l]{ 
\textbf{Transformer}: a dog holding a frisbee in the water\\
\textbf{$\mathcal{M}^{2}$ Transformer}: a dog holding a frisbee in\\ a body of water\\
\textbf{AoA Transformer}: a dog walking during a frisbee \\in a stone day\\ 
\textbf{VisualGPT (ours)}:a dog walking through the water\\ with a frisbee} 
\end{tabular}

& 
\begin{tabular}{l}{
\makecell[l]
{\textbf{GT1}:two dogs are playing on the beach \\catching a frisbee \\
\textbf{GT2}: of two dogs only one may be the victor\\
\textbf{GT3}: a dog catching a frisbee by another\\ dog on a beach\\
\textbf{GT4}: dog jumping up in the air to catch a\\ frisbee in the summer time\\
\textbf{GT5}: a dog jumping up into the air to \\catch a frisbee} }\end{tabular}
\\
\hline


\begin{tabular}{l}
     \includegraphics[width=35mm,height=32mm
]{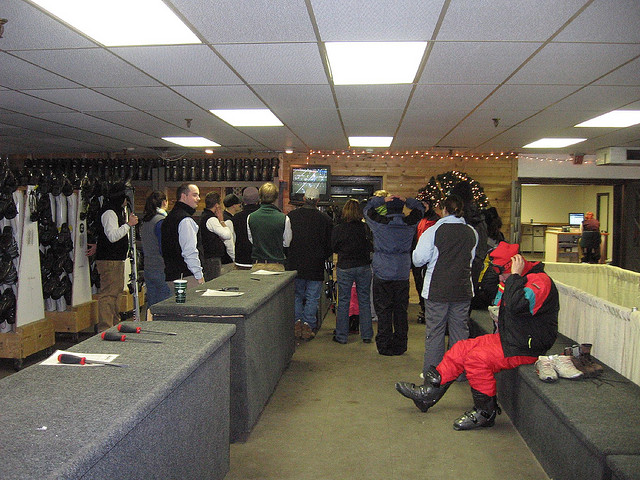} 
\end{tabular}

& 
\begin{tabular}{l}
    \makecell[l]{ 
\textbf{Transformer}: a group of people taking a\\ child in a in a building\\
\textbf{$\mathcal{M}^{2}$ Transformer}: a group of people in\\ an airport with their hands\\
\textbf{AoA Transformer}: a picture of a young \\ group of people standing for men\\ 
\textbf{VisualGPT (ours)}: a group of people\\ standing around a tv} 
\end{tabular}

& 
\begin{tabular}{l}{
\makecell[l]
{\textbf{GT1}: a group of men standing around a room \\
\textbf{GT2}: some people are waiting in a long room\\
\textbf{GT3}: people are standing in a room looking \\at a television screen\\
\textbf{GT4}: a person sitting on a bench while the\\ rest look somehwere else\\
\textbf{GT5}: a man in red winter clothes sits on\\ a bench with people behind him gather in\\ front of a tv}}\end{tabular}
\\
\hline


\begin{tabular}{l}
     \includegraphics[width=35mm,height=32mm
]{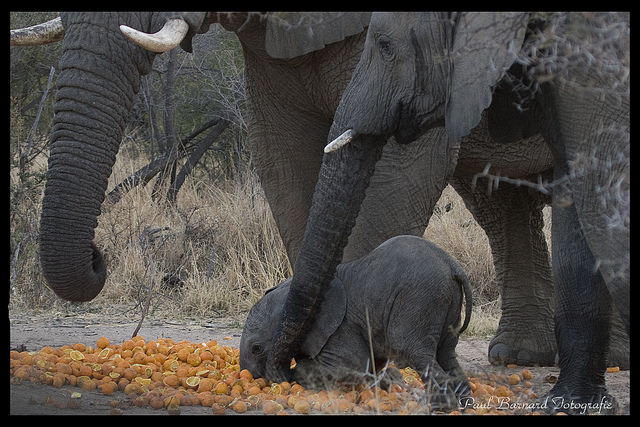} 
\end{tabular}

& 
\begin{tabular}{l}
    \makecell[l]{ 
\textbf{Transformer}: an elephant eating a elephant\\ has a elephant\\
\textbf{$\mathcal{M}^{2}$ Transformer}:  elephant with its trunk\\ with their elephant with its trunk\\
\textbf{AoA Transformer}: two elephants standing at\\ a lot of trees\\ 
\textbf{VisualGPT (ours)}: three elephants standing\\ next to some trees} 
\end{tabular}

& 
\begin{tabular}{l}{
\makecell[l]
{\textbf{GT1}: two adult elephants are surrounding\\ a baby elephant \\
\textbf{GT2}: a baby elephant kneeling in front of\\ two bigger elephants\\
\textbf{GT3}: a baby elephant and it 's parents\\ eat fruit\\
\textbf{GT4}: elephants eat fruit a baby elephant\\ rummaging in the food\\
\textbf{GT5}: a pair of adult elephants with a baby\\ elephant eat from a pile of fruit}}\end{tabular}
\\
\hline
\\
\\

\end{tabular}
\caption{Caption generated by our VisualGPT, Transformer, $\mathcal{M}^{2}$ Transformer and AoA Transformer on $0.5\%$ MS COCO data split}
\label{supp_table2}
\end{table*}

\begin{table*}

\setlength{\tabcolsep}{3pt}
\begin{tabular}{lll}
\hline

Image & Generated Captions & Ground Truth \\ \hline
\begin{tabular}{l}
     \includegraphics[width=35mm,height=32mm
]{sup_image/1.jpg} 
\end{tabular}

& 
\begin{tabular}{l}
    \makecell[l]{ 
\textbf{Transformer}: a man in a suit and a woman\\ standing in a shop \\ 
\textbf{$\mathcal{M}^{2}$ Transformer}: a man is standing in\\ a shop with a people holding people\\
\textbf{AoA Transformer}:  a man is working on a bus\\ in a \\ 
\textbf{VisualGPT (ours)}: a group of people standing\\ at an airport with their luggage} 
\end{tabular}
& 
\begin{tabular}{l}{
\makecell[l]
{\textbf{GT1}: several people are purchasing tickets\\ at a bus station \\
\textbf{GT2}: some people are checking in at the\\ ticket counter somewhere in asia\\
\textbf{GT3}: people waiting in line with luggage\\ at a ticket counter\\
\textbf{GT4}: people are standing near an airport\\ ticket kiosk\\
\textbf{GT5}: customers stand at a kiosk waiting\\ for tickets} }\end{tabular}
\\
\hline

\begin{tabular}{l}
     \includegraphics[width=35mm,height=32mm
]{sup_image/2.jpg} 
\end{tabular}

& 
\begin{tabular}{l}
    \makecell[l]{ 
\textbf{Transformer}: a bus that is parked in front\\ of a building\\
\textbf{$\mathcal{M}^{2}$ Transformer}: a couple of people walking\\ down the side of a street\\
\textbf{AoA Transformer}:  a bus is parked in a city\\ street \\ 
\textbf{VisualGPT (ours)}: a while and blue bus is\\ parked on the side of a city street} 
\end{tabular}
& 
\begin{tabular}{l}{
\makecell[l]
{\textbf{GT1}: people standing outside of a blue and\\ white bus \\
\textbf{GT2}: an image of a tour bus that is picking\\ people up\\
\textbf{GT3}: several people standing around buses\\ and most wearing orange vests\\
\textbf{GT4}: a public transit bus pulling up to pick\\ up passengers\\
\textbf{GT5}: a city bus at a stop waiting to pick up\\ passengers} }\end{tabular}
\\
\hline


\begin{tabular}{l}
     \includegraphics[width=35mm,height=32mm
]{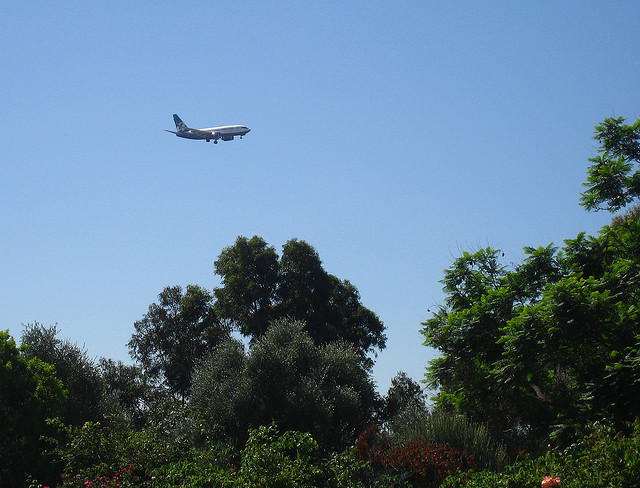} 
\end{tabular}

& 
\begin{tabular}{l}
    \makecell[l]{ 
\textbf{Transformer}: a blue and white airplane flying\\ through a sky\\
\textbf{$\mathcal{M}^{2}$ Transformer}: an air plane flying in the\\ air\\
\textbf{AoA Transformer}:  a plane airplane flying\\ down in the sky \\ 
\textbf{VisualGPT (ours)}: a plane is flying in the air\\ over the trees} 
\end{tabular}

& 
\begin{tabular}{l}{
\makecell[l]
{\textbf{GT1}: there 's and airplane in the sky flying\\ over some trees \\
\textbf{GT2}: a large plane is flying over a crowd \\of trees\\
\textbf{GT3}: a aeroplane soaring high in the sky\\ above the trees\\
\textbf{GT4}: a passenger plane flies in the sky \\over a forest\\
\textbf{GT5}: an airplane is seen flying over several\\ trees} }\end{tabular}
\\
\hline


\begin{tabular}{l}
     \includegraphics[width=35mm,height=32mm
]{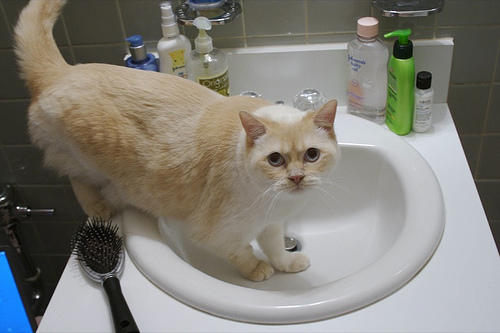} 
\end{tabular}

& 
\begin{tabular}{l}
    \makecell[l]{ 
\textbf{Transformer}: a white toilet sitting in a \\white bathroom next to a sink\\
\textbf{$\mathcal{M}^{2}$ Transformer}:  a cat sitting in the toilet\\
\textbf{AoA Transformer}:  a bathroom with a toilet\\ and a sink \\
\textbf{VisualGPT (ours)}: a cat sitting on top of a\\ bathroom sink} 
\end{tabular}

& 
\begin{tabular}{l}{
\makecell[l]
{\textbf{GT1}: a cat climbing into a bathroom sink\\ looking at someone \\
\textbf{GT2}: a cat looks up as it stands in the\\ bathroom sink\\
\textbf{GT3}: a large cat stands inside of a clean\\ bathroom sink\\
\textbf{GT4}: cat is caught stepping in to the\\ bathroom sink\\
\textbf{GT5}: a cute kitty cat in the sink of a\\ bathroom near a brush and other items}}\end{tabular}
\\
\hline


\begin{tabular}{l}
     \includegraphics[width=35mm,height=32mm
]{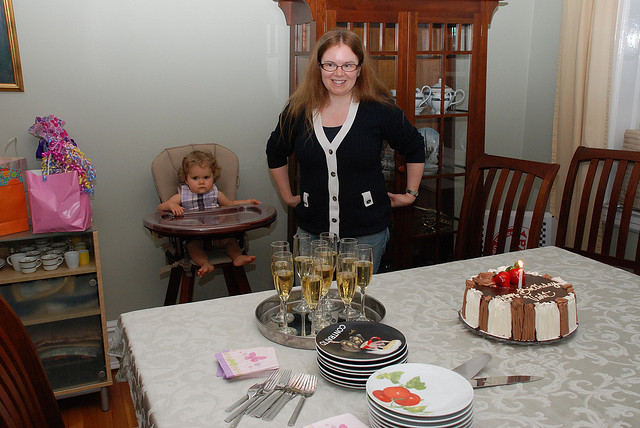} 
\end{tabular}

& 
\begin{tabular}{l}
    \makecell[l]{ 
\textbf{Transformer}: a little girl is eating a\\ birthday cake\\
\textbf{$\mathcal{M}^{2}$ Transformer}:  a child and a child are \\sitting at a table with table with table\\
\textbf{AoA Transformer}:  two children sitting at a \\table with a laptop computer\\ 
\textbf{VisualGPT (ours)}: a woman and a girl sitting\\ at a table with a birthday cake} 
\end{tabular}

& 
\begin{tabular}{l}{
\makecell[l]
{\textbf{GT1}: a woman and child stand next to a\\ table with cake on it \\
\textbf{GT2}: a lady standing near the table with a\\ baby is posing for the camera\\
\textbf{GT3}: a woman stands beside a baby in a\\ high chair a table is set with a birthday \\cake and champagne\\
\textbf{GT4}: a woman setting up her house for a \\party\\
\textbf{GT5}: a person standing next to a child in a\\ booster seat}}\end{tabular}
\\
\hline
\\

\end{tabular}
\caption{Caption generated by our VisualGPT, Transformer, $\mathcal{M}^{2}$ Transformer and AoA Transformer on $1\%$ MS COCO data split}
\label{supp_table3}
\end{table*}


\end{document}